%% file: main.tex
\documentclass{article}

 \PassOptionsToPackage{numbers,compress}{natbib}
 \usepackage[dblblindworkshop,final]{neurips_2026}

\usepackage[utf8]{inputenc} 
\usepackage[T1]{fontenc}    
\usepackage{hyperref}       
\usepackage{url}            
\usepackage{booktabs}       
\usepackage{amsfonts}       
\usepackage{nicefrac}       
\usepackage{microtype}      
\usepackage{xcolor}         
\usepackage{amsmath}        
\usepackage{algorithm}      
\usepackage{algpseudocode}  
\usepackage{multirow}
\usepackage{float}         
\floatstyle{ruled}        
\restylefloat{algorithm}
\usepackage{graphicx}
\usepackage[export]{adjustbox}   
\usepackage{xcolor}
\newcommand{\Hcal}{\mathcal{H}}
\newcommand{\Tcal}{\mathcal{T}}
\usepackage{amssymb}          
\newcommand{\jacc}{J}          
\title{Rethinking the Effectiveness of Contrastive Decoding in Mitigating Hallucinations in MLLMs}

\author{%
  Arnav Bendre \quad Guneesh Gupta \quad Shreyansh Modi \quad Kavish Grover \quad Chayan Aggarwal \\
  \vspace{3pt} \\
  Indian Institute of Technology, Roorkee \\
  \vspace{3pt} \\
  \texttt{\{arnav\_b@hre, guneesh\_g@ee, shreyansh\_m@ee, kavish\_g@me, chayan\_a@ma\}.iitr.ac.in}
}

\begin{document}

\maketitle

\begin{abstract}
Multimodal large language models frequently describe objects that are not present in an image.
Contrastive decoding has become a widely used inference-time remedy, and it is assumed to work by
contrasting a normal forward pass against a hallucination-prone one so that hallucinated content
is suppressed. However, whether the reported benchmark improvements actually arise from this
mechanism has not been examined. Our key observation is that a correction which suppresses
hallucination must depend on whether the model is hallucinating, and that this dependence can be
measured directly inside the model. Following this idea, we analyse the internal computation of
three contrastive decoding methods on both discriminative and generative tasks, and compare each
against controls that remove the contrastive term while preserving its effect on the output. The
correction turns out to be unrelated to hallucination at every layer, the gains in captioning come
instead from a constraint that narrows sampling towards greedy decoding, and an unstructured
perturbation of the same magnitude performs at least as well. Experiments across three models and
standard hallucination benchmarks show that these improvements reflect a change in decoding
behaviour rather than genuine hallucination mitigation.
\end{abstract}

\section{Introduction}
In recent years, there has been an increase in focus on models that combine vision and
text, known as Multimodal Large Language Models (MLLMs) \citep{liu2024improvedbaselinesvisualinstruction,
bai2025qwen25vltechnicalreport, zhu2025internvl3exploringadvancedtraining}. These models are typically built by connecting a
pretrained vision encoder \citep{radford2021learningtransferablevisualmodels} to a pretrained large language model
through a projection module that aligns visual features with the language model's
embedding space. Despite their rapid progress, the two modalities remain imperfectly
aligned, and MLLMs are therefore prone to hallucination, producing text that is
factually incorrect or unsupported by the input image
\citep{liu2024surveyhallucinationlargevisionlanguage}. The most studied form is object hallucination, where
the model describes objects that are not present in the image \citep{rohrbach2019objecthallucinationimagecaptioning,liu2024mitigatinghallucinationlargemultimodal}. Because these hallucinations pose a threat in high-stakes situations such
as real-world deployments, measuring and mitigating vision-language misalignment has
become an important research problem.

Contrastive Decoding is a training-free inference strategy for addressing hallucinated generations \citep{li2023contrastivedecodingopenendedtext, leng2023mitigatingobjecthallucinationslarge}. Rather than updating model parameters, contrastive decoding contrasts distributions from the original input against a hallucination-prone counterpart and suppresses token logits favored mainly by the latter \citep{leng2023mitigatingobjecthallucinationslarge, wang2024mitigatinghallucinationslargevisionlanguage}. Contrastive decoding has shown marked gains on standard hallucination benchmarks, making it an effective, computationally lightweight approach \citep{leng2023mitigatingobjecthallucinationslarge, huo2025selfintrospectivedecodingalleviatinghallucinations, lee2024delvevisualcontrastivedecoding}.

Recent work has begun to question whether contrastive decoding genuinely reduces object hallucination in
MLLMs \citep{yin2026mirageperformancegainscontrastive}. This scrutiny, however, is
purely statistical and limited to discriminative benchmarks such as POPE
\citep{li2023evaluatingobjecthallucinationlarge}. It shows that the reported gains are unreliable, but not why CD
fails to help. We extend this critique in two directions. First, we complement the
statistical evidence with a mechanistic account, tracing inside the network where
grounded and hallucinated predictions diverge. Second, we move the analysis beyond
discriminative benchmarks to free-form generation, and show that contrastive decoding is ineffective in both
settings. On this basis we argue that the benchmark gains of contrastive decoding are spurious rather than genuine mitigation of hallucination, and substantiate this through
the following findings:
\begin{itemize}
    \item \textbf{No selectivity in discriminative settings.} The contrastive shift is
    unrelated to hallucination status at every layer of the network, even where object
    presence is linearly decodable.

    \item \textbf{Gains attributable to greedy decoding.} In free-form captioning, the
    adaptive plausibility constraint drives generation towards greedy decoding, and
    greedy decoding accounts for most of the improvement reported by VCD
    \citep{leng2023mitigatingobjecthallucinationslarge}, SID
    \citep{huo2025selfintrospectivedecodingalleviatinghallucinations} and ICD
    \citep{wang2024mitigatinghallucinationslargevisionlanguage}.

    \item \textbf{No hallucination-specific signal.} The contrastive adjustment
    amplifies hallucinated object tokens as often as it suppresses them, and replacing
    it with Gaussian noise leaves benchmark scores unchanged or even better.
\end{itemize}

\section{Related Work}
\textbf{Multimodal LLMs and Object Hallucination.}
Contemporary MLLMs attach a pretrained vision encoder, typically a CLIP-style backbone \citep{radford2021learningtransferablevisualmodels}, to a generative LLM decoder, enabling joint decoding of visual and textual tokens. They differ mainly in how visual features are aligned to the language space: Q-Former-based models such as BLIP-2 \citep{li2023blip2bootstrappinglanguageimagepretraining}, MiniGPT-4 \citep{zhu2023minigpt4enhancingvisionlanguageunderstanding}, and InstructBLIP \citep{dai2023instructblipgeneralpurposevisionlanguagemodels} resample encoder outputs into a fixed-length visual prefix, while projection-based models such as LLaVA \citep{liu2024improvedbaselinesvisualinstruction}, Qwen-VL \citep{bai2023qwenvlversatilevisionlanguagemodel}, Qwen2.5-VL \citep{bai2025qwen25vltechnicalreport}, and InternVL3 \citep{zhu2025internvl3exploringadvancedtraining} use a lightweight linear layer, varying primarily in vision encoder choice and token allocation \citep{chen2024fargpt4vclosinggap, chen2024internvlscalingvisionfoundation}. In all cases, the image is passed only at input, leaving generation driven by a text-trained decoder, a design that gives rise to object hallucination, where models describe objects absent from the image \citep{rohrbach2019objecthallucinationimagecaptioning, liu2024mitigatinghallucinationlargemultimodal}, commonly attributed to statistical bias toward frequent co-occurring objects and the decoder's language prior dominating when visual evidence is weak \citep{leng2023mitigatingobjecthallucinationslarge, liu2024surveyhallucinationlargevisionlanguage}. Mitigation approaches include training-based methods that fine-tune on curated data 
\citep{liu2024mitigatinghallucinationlargemultimodal, gunjal2024detecting, sun2023aligninglargemultimodalmodels}, post-hoc verifiers that correct outputs without retraining \citep{Yin_2024}, and decoding-based methods that modify token selection at inference time. Among these methods, contrastive decoding \citep{li2023contrastivedecodingopenendedtext, leng2023mitigatingobjecthallucinationslarge, wang2024mitigatinghallucinationslargevisionlanguage} is the most widely adopted and the focus of this work.

\textbf{Contrastive Decoding Strategies.}
Contrastive decoding was first used in text-only LLMs, where a weak amateur
model is run alongside the main model and tokens that only the amateur prefers are
suppressed \citep{li2023contrastivedecodingopenendedtext, chuang2024doladecodingcontrastinglayers}. In MLLMs, the amateur is the same
model given a corrupted input, and methods differ mainly in how that input is corrupted
\citep{lee2024delvevisualcontrastivedecoding}. We focus on three widely used methods. Visual contrastive decoding \citep{leng2023mitigatingobjecthallucinationslarge}
adds noise to the image. Instruction contrastive decoding \citep{wang2024mitigatinghallucinationslargevisionlanguage} adds a misleading instruction to the
prompt. Self-Introspective decoding \citep{huo2025selfintrospectivedecodingalleviatinghallucinations} keeps only the least useful visual tokens in the early
layers of the decoder. Other methods change the corruption in similar ways, such as
removing or cropping the image \citep{favero2024multimodalhallucinationcontrolvisual, chen2024halcobjecthallucinationreduction, manevich2024mitigatinghallucinationslargevisionlanguage, zhu2024ibdalleviatinghallucinationslarge}, but the idea stays the same. All of them assume that subtracting a
hallucination-prone distribution at the final logit layer removes the cause of
hallucination, rather than merely reshaping the output. We test that assumption.

\section{Preliminaries}
An MLLM with parameters $\theta$ takes a visual input $v$ and a textual query $x$, and
generates a response one token at a time. At step $t$, the model produces a logit
$z_\theta(y_t \mid v, x, y_{<t})$ for each candidate token $y_t$, conditioned on the
image, the query, and the tokens generated so far. The next token is then sampled from
the corresponding distribution,
\begin{equation}
y_t \sim p_\theta(y_t \mid v, x, y_{<t}) = \mathrm{softmax}\big(z_\theta(y_t \mid v, x, y_{<t})\big).
\end{equation}
Hallucination arises when this distribution places high probability on objects that are
not supported by $v$.

\textbf{Contrastive decoding} attacks this imbalance without retraining. The same model is run a second time on a perturbed input
$(v^{*}, x^{*})$ designed to be more hallucination-prone, giving an \emph{amateur}
distribution in which the language prior is even more dominant. Subtracting the amateur
logits from the original \emph{expert} logits should then cancel the prior-driven
component and leave the visually grounded one,
\begin{equation}
\label{eq:cd}
z_{\mathrm{cd}}(y_t) = (1 + \alpha)\, z_\theta(y_t \mid v, x, y_{<t})
- \alpha\, z_\theta(y_t \mid v^{*}, x^{*}, y_{<t}),
\end{equation}
where $\alpha > 0$ sets the strength of the contrast. Tokens the amateur favors are
pushed down and made less probable. All variants below are
instances of Eq.~\eqref{eq:cd}; they differ only in how the amateur is built. At the commonly used setting $\alpha = 1$,
\begin{equation}
    z_{\mathrm{cd}}(y_t)\big|_{\alpha=1}
    = 2\, z_\theta(y_t \mid v, x, y_{<t}) - z_\theta(y_t \mid v^\ast, x^\ast, y_{<t})
    = z_\theta(y_t \mid v, x, y_{<t}) + d_t,
    \label{eq:cd-alpha1}
\end{equation}
where
\begin{equation}
    d_t = z_\theta(y_t \mid v, x, y_{<t}) - z_\theta(y_t \mid v^\ast, x^\ast, y_{<t})
    \label{eq:logit-gap}
\end{equation}
is the expert–amateur logit gap defined as \emph{Contrastive Adjustment} at step $t$. For notational convenience in our analysis that follows, we write
$C$ as $ z_{\mathrm{cd}}(y_t)$ for the contrastive logit,
$E$ as $ z_\theta(y_t \mid v, x, y_{<t})$ for the expert logit, 
$A$ as $z_\theta(y_t \mid v^\ast, x^\ast, y_{<t})$ for the amateur logit and Contrastive Adjustment as $d_t$ = E - A.

\textbf{Visual Contrastive Decoding} perturbs the image. It corrupts $v$ with a forward diffusion
process, $v^{*} = \sqrt{\bar{\alpha}_T}\, v + \sqrt{1 - \bar{\alpha}_T}\, \epsilon$ with
$\epsilon \sim \mathcal{N}(0, I)$, and leaves the prompt unchanged ($x^{*} = x$). The
motivation is that as visual detail is destroyed, the model has little to condition on
except its language prior and its co-occurrence statistics, so the amateur distribution
concentrates exactly on the tokens that hallucination is attributed to.

\textbf{Instruction Contrastive Decoding} perturbs the prompt instead. It prepends a misleading role
instruction to obtain $x^{*}$ while keeping the image intact ($v^{*} = v$). Such a
prefix degrades the alignment between the visual and textual streams and raises the
model's uncertainty, which is again expected to make the amateur more prone to
hallucinated objects, but through the instruction pathway rather than the visual one.

\textbf{Self-Introspective Decoding} leaves both inputs unchanged and perturbs the forward pass. Using
the attention of the early decoder layers, it ranks the visual tokens by importance and
retains only the least informative ones, producing an amateur model $\theta^{*}$ whose
visual evidence is weak but whose textual context is preserved. The second term of
Eq.~\eqref{eq:cd} then becomes $z_{\theta^{*}}(y_t \mid v, x, y_{<t})$. Because the
amateur re-runs only the remaining layers over a much shorter visual sequence, SID is
cheaper than variants that require a full second forward pass.

\textbf{Adaptive Plausibility Constraint (APC)} restricts the candidate token set to tokens that remain sufficiently likely under the original expert MLLM distribution. Its purpose is to prevent contrastive decoding from assigning high probability to implausible tokens that arise only from the logit subtraction. Following the original formulation, the head vocabulary at decoding step $t$ is
\begin{equation}
    \mathcal{V}_{\mathrm{head}}(y_{<t})
    = \left\{ y_t \in \mathcal{V}:
    p_{\theta}(y_t \mid v,x,y_{<t})
    \geq \beta \max_{w \in \mathcal{V}} p_{\theta}(w \mid v,x,y_{<t})
    \right\},
\end{equation}
where $\beta$ is the truncation threshold and $\mathcal{V}$ is the vocabulary.
APC assigns zero probability to tokens outside $\mathcal{V}_{\mathrm{head}}$, so sampling is restricted to the original model's sufficiently high-probability region.

\section{Methodology of Analysis}
We analyse contrastive decoding on both discriminative tasks and free-form generation. On
discriminative tasks such as object-existence polling in POPE, we use a logit-lens
analysis \citep{nostalgebraist2020logitlens} to locate the layer at which grounded and
hallucinated predictions start to diverge. In free-form generation, we inspect the
contrastive adjustment at each step to see whether hallucinated object tokens are
suppressed more than grounded ones.
\subsection{Discriminative Analysis}
\label{sec:disc-method}

\citet{yin2026mirageperformancegainscontrastive} argue that the reported gains of
contrastive decoding on POPE are misleading rather than genuine hallucination
mitigation. Their analysis, however, is confined to the output distribution. It
establishes that the reported gains are unreliable, but leaves open where in the model
the error is produced and how contrastive decoding acts on it. We address this question by examining the
model's internal representations.

Contrastive decoding acts on the final logits, but the prediction it modifies is built up
across the layers of the decoder. To locate where the answer is formed, we read out the
model's decision at each layer with the logit lens, which passes an intermediate hidden state through the
model's own frozen output head. We
restrict the resulting scores to the two answer tokens and take their difference,
\begin{equation}
    m_\ell \;=\; z^{\mathrm{Yes}}_\ell \;-\; z^{\mathrm{No}}_\ell ,
    \label{eq:margin}
\end{equation}
which we call the margin at layer $\ell$, the log-odds with which the model would answer
``Yes'' if decoding stopped there. Because $m_\ell$ is a difference of two logits, it is exactly the log-odds of
``Yes'' against ``No'', and every shift reported below is in the same units. Computing this margin on the expert pass, which uses
the clean input, and on the amateur pass, which uses the perturbed input, gives
$m^{E}_\ell$ and $m^{A}_\ell$. Their difference,
\begin{equation}
    \Delta_\ell \;=\; m^{E}_\ell - m^{A}_\ell ,
    \label{eq:lens-delta}
\end{equation}
is the contrastive effect at layer $\ell$. Contrastive
decoding adds this term to the expert margin, giving
$m^{\mathrm{cd}}_\ell = m^{E}_\ell + \alpha \Delta_\ell$, so $\Delta_\ell$ is negative
when the contrast suppresses ``Yes'' and positive when it reinforces it.

We restrict the analysis to questions the model answers ``Yes'', and split them by ground
truth into hallucinations $\Hcal$, where the object is absent, and correct positives
$\Tcal$, where it is present. A method that mitigates hallucination must suppress ``Yes''
on $\Hcal$ while leaving $\Tcal$ intact.Since separating $\Hcal$ from $\Tcal$ amounts to determining object presence, we train a
linear probe on the hidden state at each layer to predict presence, thereby establishing
the depth from which the information required for selective suppression is available.

Whether the contrastive effect uses this information is measured at each layer by four
quantities. We first sort all ``Yes'' questions by $-\Delta_\ell$, the extent to which
contrastive decoding pushes the answer away from ``Yes''. A selective method would place
the hallucinations $\Hcal$ at the top of this ordering. We therefore report its AUC,
which is the probability that a randomly chosen member of $\Hcal$ is ranked above a
randomly chosen member of $\Tcal$ and equals $0.5$ when the shift is unrelated to
hallucination. We also report the fraction of $\Hcal$ for which $\Delta_\ell > 0$. These
are the hallucinated answers that contrastive decoding pushes further towards ``Yes''.
The third quantity is $R^2_\ell$, the share of the variance of $\Delta_\ell$ that
hallucination status explains. The fourth is the accuracy of the presence probe on the
amateur pass. A presence signal retained by the amateur cancels under subtraction and
cannot appear in $\Delta_\ell$. Section~\ref{sec:disc-results} reports all four
quantities, with the layer-wise curves in Figure~\ref{fig:disc-battery} and the
supporting values in Appendix~\ref{app:disc-depth}.

A negative result may also arise from an insensitive measurement. We therefore report two
controls. The first establishes the sensitivity of the test. We construct an oracle shift
that suppresses ``Yes'' on $\Hcal$ by design and add it to $\Delta$ at progressively
smaller strengths. The smallest strength that the test still detects gives the size of an
effect that could not have been missed. Appendix~\ref{app:disc-oracle} gives the
construction of the oracle shift and reports this value for all three methods. The second control establishes whether the per-question
values of $\Delta$ are required. We replace $\Delta$ with a single constant applied to
all $9{,}000$ questions and sweep this constant against POPE accuracy. Such a constant
carries no information about any image. If it reproduces the accuracy obtained with the
true $\Delta$ then the per-question values contribute nothing, as shown in
Appendix~\ref{app:scalar-bias}.

\subsection{Generative Analysis}
In the generative setting we first measure how far the plausibility constraint drives
generation towards greedy decoding and whether the reported gains can be attributed to
this shift. We then examine whether contrastive decoding suppresses
hallucinated tokens in any meaningful way.
\subsubsection{Spurious Gains via Contrastive Decoding}
\label{4.2.1}
Contrastive decoding is paired with an adaptive plausibility constraint, which restricts
the candidate set to tokens assigned sufficiently high probability by the expert model
before the contrast is applied. To isolate the effect of this constraint on the
generated text, we produce three answers with the same model: one by greedy decoding,
one by direct sampling without the constraint, and one by sampling with the constraint
applied at threshold $\beta$. Algorithm~\ref{alg:sample-apc} states this third procedure, which we denote as
\emph{APC-S}. We then measure how similar the constrained answer is to
each of the other two, and average over all questions. Similarity is
measured by the Jaccard score over the unique sub-word tokens of two answers, defined in
Eq.~\eqref{eq:jaccard}. We repeat this at values of $\beta$ from 0 to 1. If the
constraint reduces sampling to greedy search, the similarity of constrained sampling to
greedy decoding should rise as $\beta$ increases, while its similarity to direct
sampling stays low and flat. Because this score ignores word order and counts common
function words, we repeat the sweep with two stricter variants, one over bigrams and one
over content words alone, and report both in Appendix~\ref{app:jaccard-checks}.

Figure ~\ref{fig:apc-both-refs} shows that constrained sampling becomes
increasingly similar to greedy decoding as $\beta$ grows. We therefore test whether the
improvements attributed to contrastive decoding follow from this resemblance rather than
from the contrast itself, evaluating greedy decoding, plain sampling, \emph{APC-S},
and contrastive decoding on CHAIR \citep{rohrbach2019objecthallucinationimagecaptioning} and LLaVA-Bench \citep{liu2023visualinstructiontuning}. Results are reported in
Tables~\ref{tab:llava_bench_sample} and~\ref{tab:chair-beta01} and analysed in
Section~\ref{sec:greedy-gains}.

\begin{algorithm}[t]
\small
\caption{\emph{APC-S}: sampling under the adaptive plausibility constraint, without a
contrastive term.}
\label{alg:sample-apc}
\begin{algorithmic}[1]
\Require Model $z_\theta$, image $v$, prompt $x$,
threshold $\beta$, temperature $\tau$, maximum length $T$
\Ensure Response $y$
\State $y \gets \langle\,\rangle$
\For{$t = 1$ \textbf{to} $T$}
  \State $z_t \gets z_\theta(\cdot \mid v, x, y_{<t})$
    \Comment{logit vector over the vocabulary}
  \State $p_t \gets \mathrm{softmax}(z_t)$
    \Comment{expert distribution, before temperature}
  \State $\mathcal{V}_t \gets \{\, w : p_t(w) \ge \beta \max_{w'} p_t(w') \,\}$
    \Comment{adaptive plausibility constraint}
  \State $\tilde{z}_t \gets z_t$, \quad $\tilde{z}_t(w) \gets -\infty$ \textbf{for all} $w \notin \mathcal{V}_t$
    \Comment{mask implausible tokens}
  \State $y_t \sim \mathrm{softmax}(\tilde{z}_t / \tau)$
    \Comment{sample from the constrained distribution}
  \State $y \gets y \Vert y_t$
  \If{$y_t = \texttt{[EOS]}$}
    \State \textbf{break}
  \EndIf
\EndFor
\State \Return $y$
\end{algorithmic}
\end{algorithm}

\subsubsection{Analysis of Semantically Meaningful Mitigation}
If contrastive decoding mitigates hallucination, its adjustment must act selectively,
suppressing hallucinated object tokens while leaving grounded ones unchanged or
amplified. We examine whether this holds by measuring the sign and magnitude of the
contrastive adjustment $d_t$ in Eq.~\eqref{eq:logit-gap}, the quantity added to the
expert logit of each candidate token. During caption generation we inspect the ten highest-scoring
candidates at every step and keep those naming an object in the vocabulary CHAIR uses
for detection. Each retained candidate is labelled \emph{real} if the ground-truth
annotations of the image contain that object, and \emph{hallucinated} otherwise. Since
the expert and amateur logits are read at the same step for the same candidate, every
$d_t$ is a paired measurement, so the two groups remain comparable once pooled across
steps and images. For each model and method we report the percentage of candidates amplified ($d_t > 0$),
the percentage suppressed ($d_t < 0$), and the mean adjustment. These statistics are
presented in Table~\ref{tab:emitted} for the object tokens emitted by the model, and in
Appendix~\ref{top10_section} for the complete set of top-10 candidates.

If the specific structure of $d_t$ is responsible for the observed improvements, then
substituting a structureless perturbation of comparable magnitude should eliminate them.
Accordingly, we replace $d_t$ with
$d_{\mathrm{proxy}} \sim \mathcal{N}(\mu, \sigma^2)$, sampled independently at each
generation step, which yields the decoding score $E + d_{\mathrm{proxy}}$. The draws are
truncated to the range observed for $d_t$, and the adaptive plausibility constraint is
applied unchanged.The resulting captions are evaluated with CHAIR, with the outcomes reported in
Table~\ref{tab:proxy-chair-perrun}.
\section{Results}

\subsection{The contrastive shift is not selective to hallucination}
\label{sec:disc-results}

\begin{figure}[!tb]
  \centering
  \includegraphics[width=0.86\linewidth]{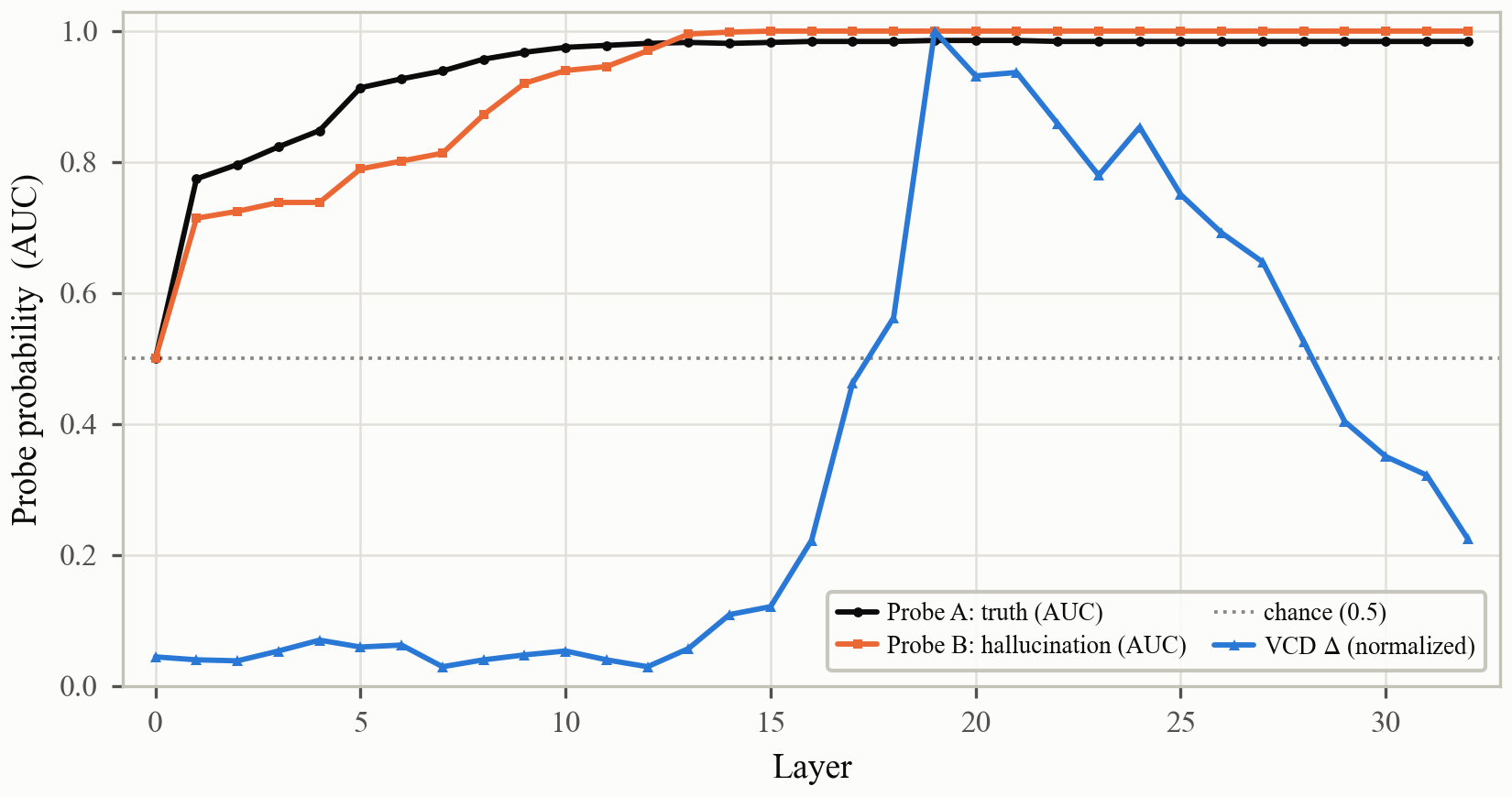}
\caption{Probe accuracy and mean contrastive shift by layer, LLaVA-v1.5-7B with VCD.
The probes saturate by layer 12, while the shift peaks at layer 19, after the decision
has been made.}
  \label{fig:probe-vs-delta}
\end{figure}

Figure~\ref{fig:probe-vs-delta} shows when the decision is formed and when the
contrastive shift arrives. Object presence becomes decodable by layer 8 and the erroneous
``Yes'' by layer 12, so the answer is settled in the middle of the network, whereas the
mean shift on $\Hcal$ stays close to zero across that range and peaks seven layers later.
Its sign is positive at most depths, so contrastive decoding moves hallucinated answers
further towards ``Yes'' rather than suppressing them. Appendix~\ref{app:disc-depth}
reports the corresponding values and decomposes the shift by block.

\begin{figure}[b]
  \centering
  \includegraphics[width=0.48\linewidth]{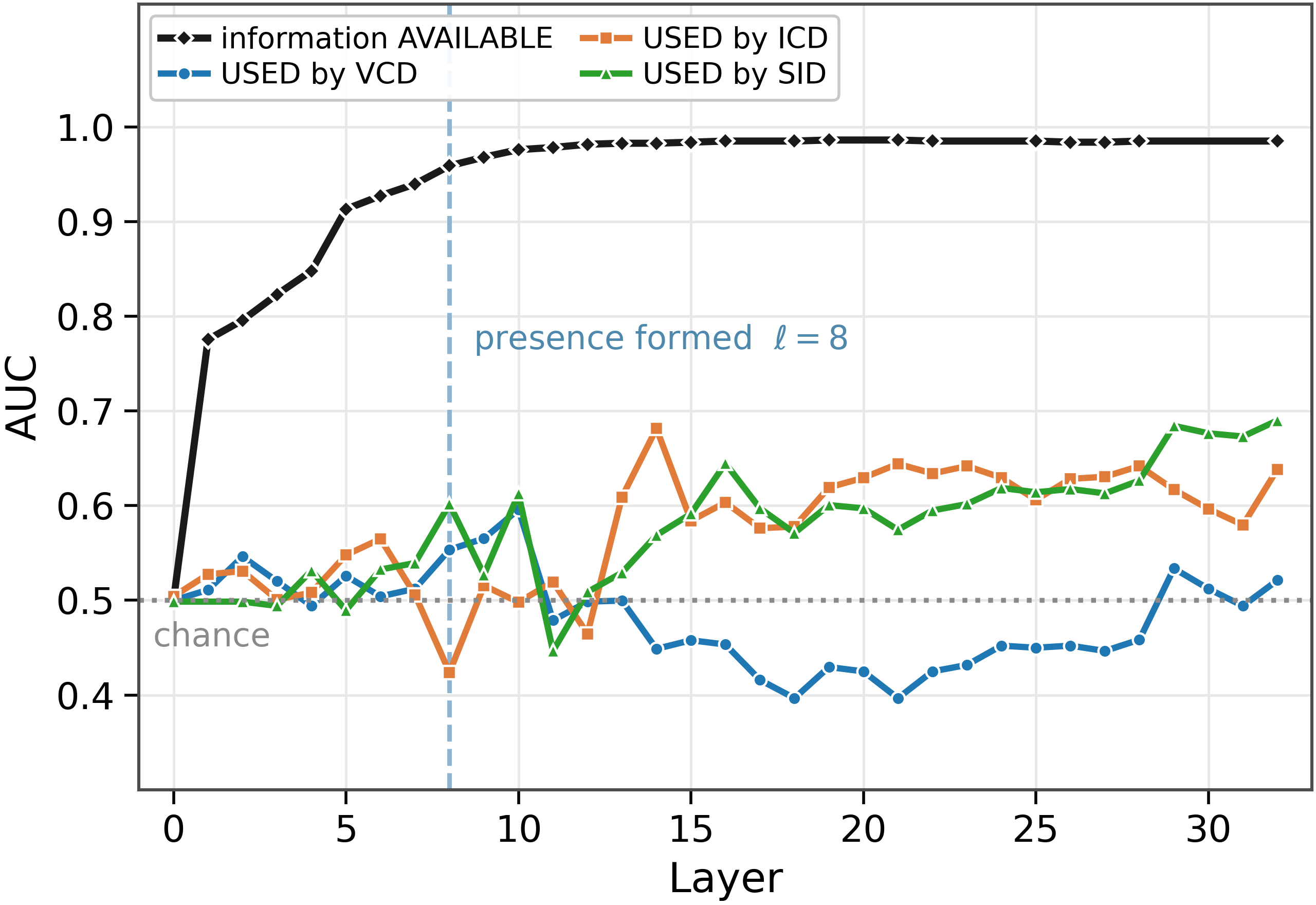}
  \hfill
  \includegraphics[width=0.48\linewidth]{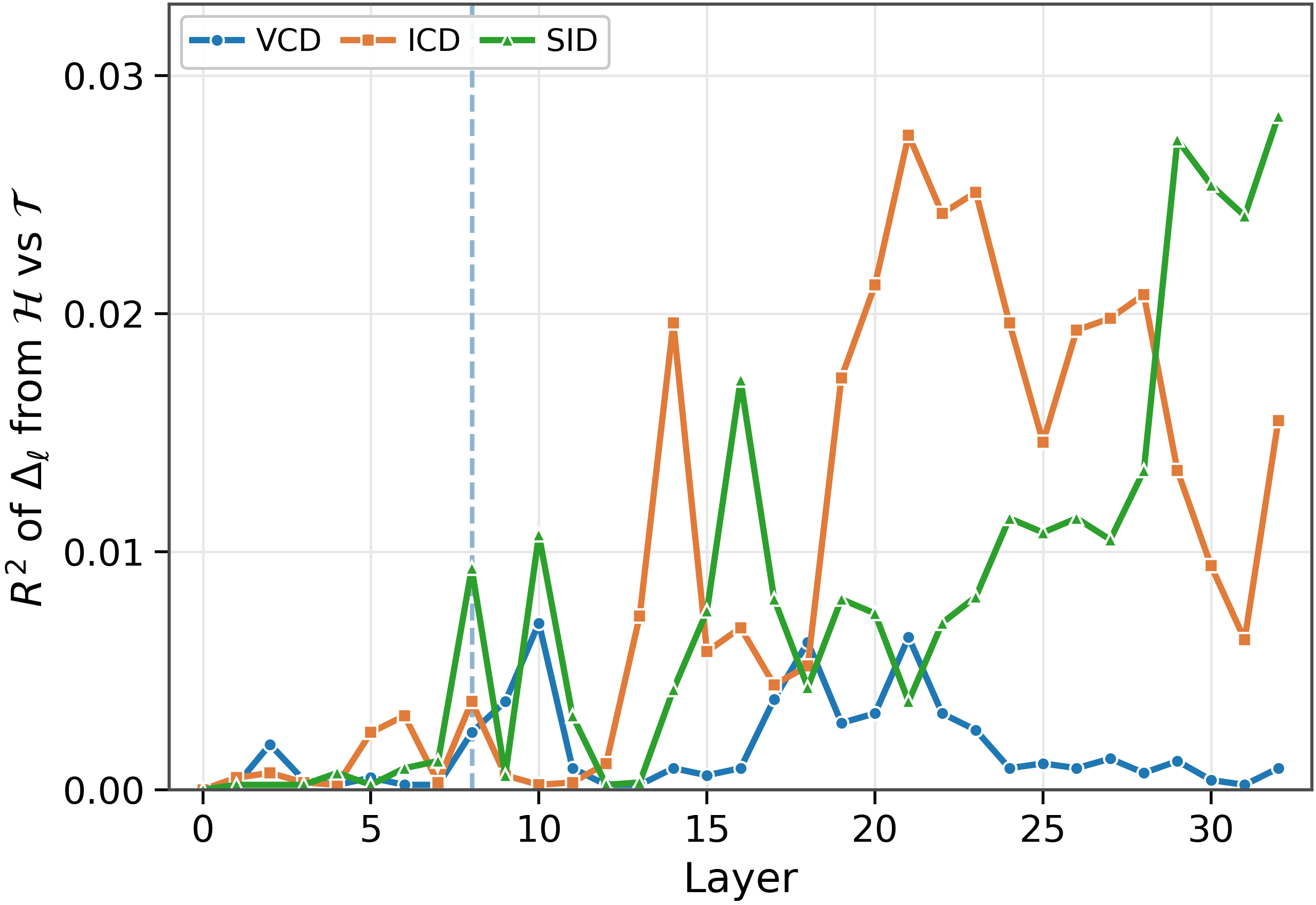}
\caption{LLaVA-v1.5-7B on POPE-COCO. \textbf{Left:} object presence is decodable
  from layer 8, yet no method's shift separates hallucinations from correct answers.
  \textbf{Right:} hallucination status explains almost none of the variance of
  $\Delta_\ell$ at any depth.}
  \label{fig:disc-battery}
\end{figure}

A selective correction requires knowing whether the named object is actually present in
the image, since this is what distinguishes a hallucinated ``Yes'' from a correct one.
Figure~\ref{fig:disc-battery} shows that this information is available inside the model:
the presence probe recovers it from layer 8 onwards, well before the answer is produced.
The contrastive shift, however, does not use it.
Table~\ref{tab:disc-battery} reports the AUC with which each shift separates $\Hcal$ from
$\Tcal$ at the output layer. For VCD this value is $0.523$, and its $95\%$ confidence
interval includes $0.5$, so the shift is unrelated to hallucination status. ICD and SID
reach $0.639$ and $0.691$, but $R^2_\ell$ in Figure~\ref{fig:disc-battery} shows that
hallucination status explains under $1\%$ of the variation in VCD's shift and under $3\%$
for ICD and SID at every depth. The relationship is therefore too weak to be useful.

To reduce hallucination, a method must lower the score of ``Yes'' when the object is
absent, that is, produce a negative mean shift $\overline{\Delta}$ on $\Hcal$. The
selectivity AUC in Table~\ref{tab:disc-battery} does not test this, since it only
compares the two groups and ignores the sign of the shift. The mean shifts in Table~\ref{tab:disc-battery} show what the ranking hides. For SID the mean shift is $+1.61$ log-odds on $\Hcal$ and $+3.11$ on $\Tcal$. Both values
are positive, so SID pushes both groups towards ``Yes''. It scores above chance only
because the push on hallucinations is the smaller of the two, not because it suppresses
them. ICD is the only method with a negative mean shift on $\Hcal$, at $-0.17$, but its shifts
on the two groups differ by only $0.11$ log-odds. The expert margin, that is the distance
of the model's answer from the boundary between ``Yes'' and ``No'', has a median of
$3.16$, so a difference of this size cannot change an answer. Neither method
performs the correction its AUC suggests.

\begin{table}[!tb]
\centering
\small
\setlength{\tabcolsep}{7pt}
\renewcommand{\arraystretch}{1.2}
\caption{Selectivity of the contrastive shift at the output layer, LLaVA-v1.5-7B
on POPE-COCO. No method separates hallucinations from correct answers.
Here $|\Hcal|=257$ for VCD and $261$ for ICD and SID, with $|\Tcal|=3453$.}
\label{tab:disc-battery}
\begin{tabular}{@{}lrcrrcc@{}}
\toprule
\multirow{2.5}{*}{Method} & \multicolumn{2}{c}{Selectivity}
 & \multicolumn{2}{c}{Mean shift $\overline{\Delta}$}
 & \multirow{2.5}{*}{Amateur AUC} & \multirow{2.5}{*}{\shortstack{\% of $\Hcal$ shifted\\towards ``Yes''}} \\
\cmidrule(lr){2-3} \cmidrule(lr){4-5}
 & AUC & 95\% CI & $\Hcal$ & $\Tcal$ & & \\
\midrule
VCD & 0.523 & [0.49,\, 0.55] & $+3.32$ & $+3.60$ & 0.777 & 96.5\% \\
ICD & 0.639 & [0.60,\, 0.67] & $-0.17$ & $-0.06$ & 0.940 & 25.7\% \\
SID & 0.691 & [0.66,\, 0.72] & $+1.61$ & $+3.11$         & 0.799 & 79.7\% \\
\bottomrule
\end{tabular}
\end{table}

The reason for this behaviour is visible in the amateur AUC column of
Table~\ref{tab:disc-battery}. Every amateur still encodes object presence almost as well
as the expert. This is clearest for ICD, whose amateur matches the expert because it
perturbs only the instruction and leaves the image untouched.
Figure~\ref{fig:disc-battery-appx} shows that the same pattern holds at every layer. The
presence signal is therefore available on both sides of the subtraction in
Eq.~\eqref{eq:lens-delta} and largely cancels. What remains in $\Delta_\ell$ carries
almost no information about whether the object is present.

Two controls confirm that this null reflects an absence rather than an insensitive test.
An oracle shift that is selective by construction reaches an AUC of $0.88$ under the same
computation, and the smallest injected selective component the test detects is between
$9$ and $24$ times smaller than the methods' own shifts, so a genuine effect of
comparable size would have been registered. Replacing the contrastive term with a single
constant added to every question is likewise sufficient: the best constant reaches
$86.86\%$ POPE accuracy against $85.49\%$ for greedy decoding, exceeding all three
contrastive methods. The per-question values of $\Delta$ therefore contribute nothing
that a constant does not. Appendices~\ref{app:disc-oracle} and~\ref{app:scalar-bias} give
full details.

Taken together, these results show that contrastive decoding does not correct
hallucination at the point where the answer is decided. It shifts the Yes/No margin by an
amount whose direction is fixed by how the amateur is constructed, not by whether the
object is present, so any change in POPE accuracy reflects a recalibration of the
decision threshold. Scaling the shift cannot repair this, since a larger $\alpha$ only
magnifies a shift that already points the wrong way on $96.5\%$ of hallucinations. The
unidirectional shift reported by \citet{yin2026mirageperformancegainscontrastive} is
therefore a uniform translation of the margin, applied without regard to the error it is
meant to remove.

\subsection{APC shifts generation towards greedy decoding}
\label{sec:apc-greedy}

\begin{figure}[b]
\centering
\includegraphics[width=\textwidth]{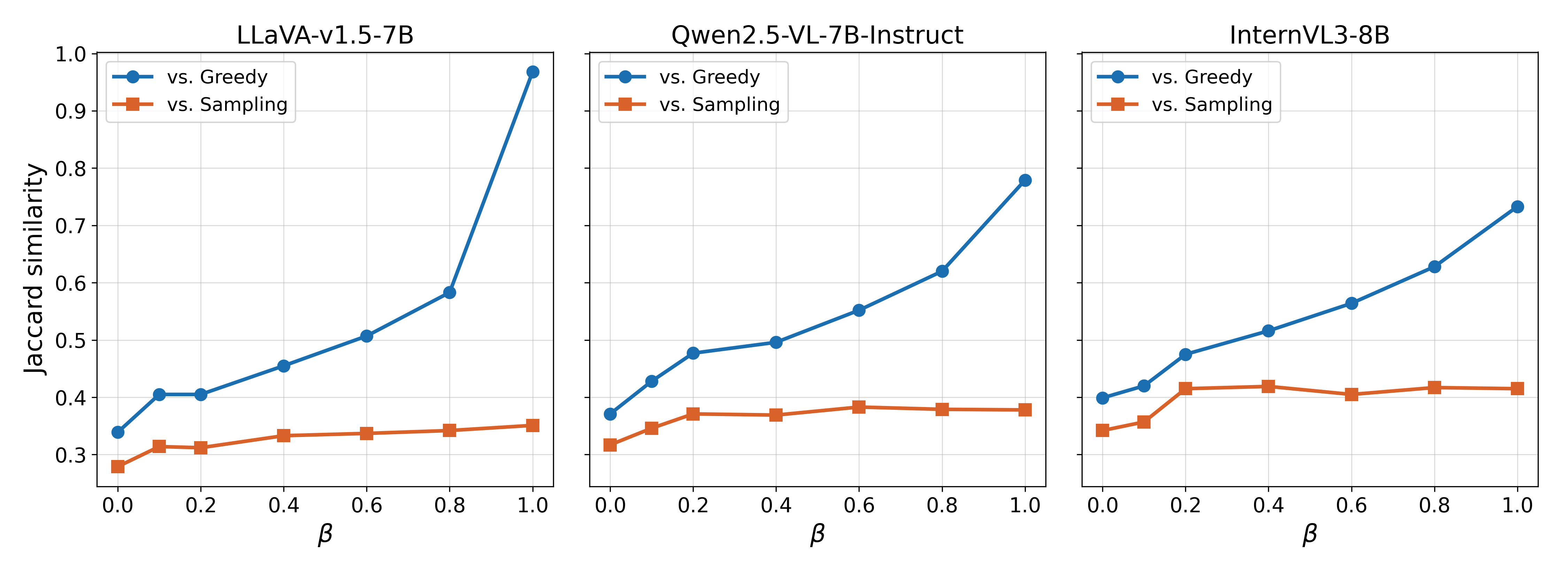}
\caption{Jaccard similarity between APC-S and greedy/direct sampling across $\beta$ thresholds on LLaVA-Bench (In-the-Wild)}
\label{fig:apc-both-refs}
\end{figure}

Figure~\ref{fig:apc-both-refs} reports the mean Jaccard similarity of APC-S to greedy
decoding and to direct sampling, over the 60 LLaVA-Bench (In-the-Wild) questions. On all
three models the two curves separate as $\beta$ increases: similarity to greedy decoding
rises steadily, while similarity to direct sampling stays flat across the entire range.
The separation is widest on LLaVA-v1.5-7B, where the constrained output becomes almost
identical to greedy decoding at the upper end of the range, but the same trend holds on
Qwen2.5-VL-7B-Instruct and InternVL3-8B. The bootstrap confidence intervals in
Appendix~\ref{sec:appendix-C-CI-Results} confirm that the $\beta = 0$ and $\beta = 1$
endpoints are disjoint. The effect is not confined to extreme thresholds: at the default
$\beta = 0.1$ used by VCD, APC-S is already closer to greedy decoding than to direct
sampling.

As $\beta$ increases the constraint alone replaces sampling with greedy decoding, so
methods that apply it at small but non-zero $\beta$ may owe their improvements over a
sampling baseline to this replacement rather than to the contrastive term, a possibility
we examine next.

\subsection{Greedy decoding accounts for the reported gains}
\label{sec:greedy-gains}
Tables~\ref{tab:llava_bench_sample} and~\ref{tab:chair-beta01} show that the reported
gains of contrastive decoding follow from greedy decoding. No contrastive method matches
plain greedy decoding on any model in CHAIR results, and APC-S attains a comparable improvement without
any contrastive term. Raising $\beta$, which moves the output closer to greedy decoding,
improves both metrics further. The reported benefit therefore reflects this shift rather
than the contrast itself.
\begin{table}[h]
\centering
\caption{LLaVA-Bench (In-the-Wild) relative scores for plain sampling and
APC-S. Results show that greedy decoding performs best on all three
models. Higher is better.}
\label{tab:llava_bench_sample}
\begin{tabular}{lccc}
\toprule
Method & LLaVA-v1.5-7B & Qwen2.5-VL-7B-Instruct & InternVL3-8B \\
\midrule
Sample & 56.1 & 73.9 & 76.3 \\
APC-S ($\beta=0.1$) & 56.7 & 80.6 & 77.1 \\
\textbf{Greedy} & \textbf{62.1} & \textbf{83.6} & \textbf{79.3} \\
\bottomrule
\end{tabular}
\end{table}

\begin{table}[H]
\centering
\caption{CHAIR scores on MS~COCO for sampling, contrastive decoding, APC-S, and greedy
decoding, across three models. Greedy decoding performs best throughout. Lower is better.}
\label{tab:chair-beta01}
\small
\setlength{\tabcolsep}{3pt}
\begin{tabular}{@{}lcccccc@{}}
\toprule
\multirow{2}{*}{\raisebox{-3pt}{Condition}} & \multicolumn{2}{c}{LLaVA-v1.5-7B} & \multicolumn{2}{c}{Qwen2.5-VL-7B-Instruct}
& \multicolumn{2}{c}{InternVL3-8B} \\
\cmidrule(lr){2-3} \cmidrule(lr){4-5} \cmidrule(lr){6-7}
 & CHAIR-S ($\downarrow$) & CHAIR-I ($\downarrow$) & CHAIR-S ($\downarrow$) & CHAIR-I ($\downarrow$) & CHAIR-S ($\downarrow$) & CHAIR-I ($\downarrow$) \\
\midrule
Sample & 57.4 & 17.3 & 31.2 & 10.4 & 33.5 & 9.1 \\
VCD ($\beta=0.1$) & 57.2 & 17.4 & 28.8 & 8.9 & 33.5 & 9.5 \\
SID ($\beta=0.1$) & 51.8 & 15.0 & 29.6 & 9.9 & 33.2 & 9.3 \\
VCD ($\beta=0.2$) & 56.2 & 16.5 & 25.1 & 8.1 & 33.1 & 9.1 \\
SID ($\beta=0.2$) & 51.5 & 14.8 & 27.8 & 8.6 & 33.0 & 9.2 \\
APC-S ($\beta=0.1$) & 53.4 & 16.3 & 23.5 & 7.5 & 29.7 & 8.4 \\
\textbf{Greedy} & \textbf{50.0} & \textbf{13.9} & \textbf{22.8} & \textbf{7.3} & \textbf{29.4} & \textbf{8.1} \\
\bottomrule
\end{tabular}
\end{table}

\subsection{Contrastive adjustment carries no hallucination signal}
\label{sec:no-signal}

Table~\ref{tab:emitted} shows that hallucinated object tokens are amplified more often
than they are suppressed, for every method and on every model. The contrastive adjustment
therefore acts in the direction opposite to the one its mechanism assumes, and it does
not distinguish hallucinated tokens from grounded ones, which are amplified at a similar
rate. These results provide no evidence that contrastive decoding selectively suppresses
hallucinated content.

Table~\ref{tab:proxy-chair-perrun} reports CHAIR scores for the Gaussian proxy and for
the contrastive methods, each averaged over three seeded runs so that the comparison
does not depend on a single noise draw. Replacing $d_t$ with a structureless
perturbation does not degrade performance, and the proxy in fact scores better than the
contrastive methods. The contrastive adjustment therefore carries no
hallucination-specific signal, since the improvement attributed to it is obtained
without any information about the image.

\begin{table}[h]
    \centering
    \caption{Percentage of emitted real and hallucinated object tokens amplified (Amp.) and
suppressed (Sup.) by the contrastive adjustment. Larger value per row is shown in bold. For every method, hallucinated tokens are
amplified more often than they are suppressed.}
    \label{tab:emitted}
    \resizebox{\textwidth}{!}{%
    \begin{tabular}{llrrrrrr}
        \toprule
        \multirow{2}{*}{\raisebox{-3pt}{Model}} & & \multicolumn{3}{c}{Real} & \multicolumn{3}{c}{Hallucinated} \\
        \cmidrule(lr){3-5} \cmidrule(lr){6-8}
        & Method & Mean $d_t$ & Amp. \% & Sup. \% & Mean $d_t$ & Amp. \% & Sup. \% \\
        \midrule
        \multirow{3}{*}{LLaVA-v1.5-7B}
            & VCD & $0.210$  & $\textbf{61.1\%}$ & $38.9\%$ & $0.451$  & $\textbf{66.5\%}$ & $33.5\%$ \\
            & SID & $0.845$  & $\textbf{60.8\%}$ & $39.2\%$ & $0.270$  & $\textbf{51.1\%}$ & $48.9\%$ \\
            & ICD & $0.091$  & $\textbf{59.9\%}$ & $40.1\%$ & $0.123$  & $\textbf{69.0\%}$ & $31.0\%$ \\
        \midrule
        \multirow{3}{*}{Qwen2.5-VL-7B-Instruct}
            & VCD & $-0.343$ & $37.5\%$ & $\textbf{62.5\%}$ & $0.026$  & $\textbf{51.1\%}$ & $48.9\%$ \\
            & SID & $1.212$  & $\textbf{70.5\%}$ & $29.5\%$ & $0.662$  & $\textbf{61.3\%}$ & $38.7\%$ \\
            & ICD & $0.221$  & $\textbf{55.1\%}$ & $44.9\%$ & $0.182$  & $\textbf{53.9\%}$ & $46.1\%$ \\
        \midrule
        \multirow{3}{*}{InternVL3-8B}
            & VCD & $0.327$  & $\textbf{54.7\%}$ & $45.3\%$ & $0.798$  & $\textbf{68.3\%}$ & $31.7\%$ \\
            & SID & $2.865$  & $\textbf{90.6\%}$ & $9.4\%$  & $2.458$  & $\textbf{87.6\%}$ & $12.4\%$ \\
            & ICD & $0.138$  & $\textbf{51.8\%}$ & $48.2\%$ & $0.324$  & $\textbf{61.5\%}$ & $38.5\%$ \\
        \bottomrule
    \end{tabular}%
    }
\end{table}

\begin{table}[h]
\centering
\caption{CHAIR scores on LLaVA-v1.5-7B. Each contrastive method is paired with a
Gaussian proxy matched to its $d_t$ statistics. Lower
is better.}
\label{tab:proxy-chair-perrun}
\small
\setlength{\tabcolsep}{8pt}
\renewcommand{\arraystretch}{1.25}
\begin{tabular}{@{}lrr@{}}
\toprule
Method & CHAIR-S ($\downarrow$) & CHAIR-I ($\downarrow$) \\
\midrule
VCD ($\beta{=}0.1$)                              & 57.2          & 17.4 \\
Proxy $\mathcal{N}(0.026,\,0.380^{2})$     & \textbf{51.4} & \textbf{14.1} \\
\addlinespace[1pt]
\midrule
\addlinespace[1pt]
SID ($\beta{=}0.1$)                              & 51.8          & 15.0 \\
Proxy $\mathcal{N}(0.088,\,0.851^{2})$     & \textbf{51.5} & 15.0 \\
\bottomrule
\end{tabular}
\end{table}

\section{Limitations and Future Work}
Computational resources bound the scope of this study. Our mechanistic
analysis covers a single model on a single discriminative benchmark because the logit-lens
procedure requires storing per-layer representations for both passes of every question, so whether
the same absence of a hallucination-specific signal holds in larger models or in free-form
generation remains open. Two properties of the evaluation metrics further qualify the generative
results. The CHAIR
metric recognises only the eighty MS~COCO object categories and resolves
alternative wordings through a fixed synonym list, so hallucinated objects outside this vocabulary
go uncounted, and the LLaVA-Bench scores come from a single language-model judge rather
than from agreement between independent judges. Both bound the absolute values without affecting the
comparisons, since every condition is scored under the same metric.

The most direct extension of this work is a mechanistic account of free-form generation, where the
layer-wise analysis must accommodate a decision distributed across many decoding
steps rather than concentrated in a single answer token. Applying the same treatment to newer methods and larger
models would establish how far these conclusions generalise.

\section{Conclusion}
This paper examined whether contrastive decoding reduces object hallucination in multimodal large
language models. We find that it does not. The contrastive shift carries no information about
whether a question is a hallucination, and this holds at every layer of the network, including
the layers at which object presence is already available to the model. What the shift does instead
is move the decision threshold, which a single constant reproduces and a structureless
perturbation matches. In free-form generation the improvement comes from the adaptive
plausibility constraint, which narrows sampling towards greedy decoding, and not from the
contrastive term at all. The gains reported on standard hallucination benchmarks are therefore a
change in decoding behaviour rather than evidence of hallucination mitigation, and inference-time
methods should be validated against greedy decoding and structureless controls before their gains
are accepted.

\bibliographystyle{unsrtnat}
\bibliography{bibliography}


\appendix
\newpage

\begin{center}
  {\LARGE\bfseries Appendix\par}
\end{center}
\vspace{0.5em}

\section{Broader Impact and Ethical Considerations}
The hallucination problem in MLLMs refers to the generation of outputs that are factually incorrect or misaligned with the visual input. Because these outputs are fluent, confident, and syntactically indistinguishable from grounded ones, they carry no surface signal of their own unreliability, making them difficult for downstream users to detect without independently verifying the source image. This becomes particularly consequential in high-stakes decision loops such as medical report generation, insurance and disaster-damage assessment, autonomous driving, and judicial evidence analysis, domains where undetected errors can have serious real-world consequences. 

On the positive side, our work advances the understanding of why existing mitigation strategies fall short, which is a necessary step toward developing genuinely effective solutions. By demonstrating that contrastive decoding methods do not meaningfully reduce object hallucinations, we caution against their uncritical deployment in safety-sensitive applications and redirect attention toward approaches that address hallucination at its root cause. We hope this encourages the community to adopt more rigorous evaluation standards when assessing hallucination mitigation methods, particularly beyond discriminative benchmarks, before such methods are considered reliable enough for real-world use. 

\section{Additional Experimentation Details}
We evaluate three models, LLaVA-v1.5-7B, Qwen2.5-VL-7B-Instruct and InternVL3-8B, on three
contrastive decoding methods, VCD, SID and ICD. Each method is run with its official
implementation. The contrastive strength is fixed at $\alpha=1$ and the adaptive plausibility
threshold at $\beta=0.1$ unless stated otherwise. The Jaccard analysis instead sweeps $\beta$
over six values in $[0,1]$.

The discriminative experiments use POPE over the COCO, A-OKVQA and GQA sources and the random,
popular and adversarial splits. The generative experiments use CHAIR on 500 images sampled from
the MS COCO validation set and LLaVA-Bench (In-the-Wild). Generation is capped at 1024 tokens
with sampling temperature $\tau=1.0$. CHAIR uses the prompt ``Describe this image in detail.''
while the LLaVA-Bench and Jaccard analyses use the LLaVA-Bench questions.

The logit-lens analysis uses LLaVA-v1.5-7B on POPE MS~COCO
\citep{lin2015microsoftcococommonobjects}, with $9{,}000$ questions pooled over the three splits.
Each question is run twice, once through the expert branch and once through the amateur branch.
The logit-lens margin is recorded at the final prompt-token position for all 33 layers. The probes
are logistic regressions fitted on standardized hidden states with 5-fold cross-validation, and
every reported score is computed from out-of-fold predictions.

The Gaussian proxy takes its mean and standard deviation from the top-10 candidate logits recorded
during the VCD and SID runs, which give about $1.68$ million and $1.49$ million values
respectively. The Jaccard analysis compares each condition with a reference, either greedy decoding
or direct sampling, and averages the per-question sub-word Jaccard index over the 60 LLaVA-Bench
questions. Two stricter variants are also reported. All confidence intervals are image-level
cluster bootstraps at the $95\%$ level with $B=10{,}000$ resamples.

LLaVA-v1.5-7B and InternVL3-8B run with full eager attention in fp16 and bf16 respectively.
Qwen2.5-VL-7B-Instruct runs in fp16 and switches to eager attention only where SID requires it.
All experiments are conducted on a single NVIDIA L4 GPU.

\section{Logit-Lens Analysis}
\label{app:disc}

This appendix supports the logit-lens analysis of Section~\ref{sec:disc-results}. It reports the
three experiments that the section states in prose: the depth comparison that motivated the
analysis, the oracle calibration that bounds what the selectivity metric can detect, and the
scalar-bias sweep.

All three use LLaVA-v1.5-7B on POPE-COCO, with $9{,}000$ questions drawn equally from the random,
popular and adversarial splits. Half of the questions name an object that is present and half name
one that is absent. Decoding is greedy with $\alpha=1$, and each condition is extracted in a single
forward pass, so the measurements are deterministic and are not averaged over seeds. The
hallucination set $\Hcal$ holds $257$ questions for VCD and $261$ for ICD and SID, against
$|\Tcal|=3453$ correct positives, so the model hallucinates on $5.7\%$ of absent-object questions.
The median expert margin on $\Hcal\cup\Tcal$ is $3.16$ log-odds, and every shift reported below
should be read against it.

Because the logit lens is used to read a decision the model has already made, it must agree with
that decision rather than merely correlate with it. Taking the sign of the output-layer margin
$m^{E}_{32}$ as a predicted answer reproduces the recorded greedy answer on all $9{,}000$ questions
for the ICD and SID extractions and on $8{,}998$ for VCD. The expert pass is not bitwise
reproducible across extractions, which also explains the $0.02$-point spread in the expert column
of Table~\ref{tab:app-scalar}.

\subsection{Depth Comparison}
\label{app:disc-depth}

\begin{figure}[h]
  \centering
  \includegraphics[width=0.48\linewidth]{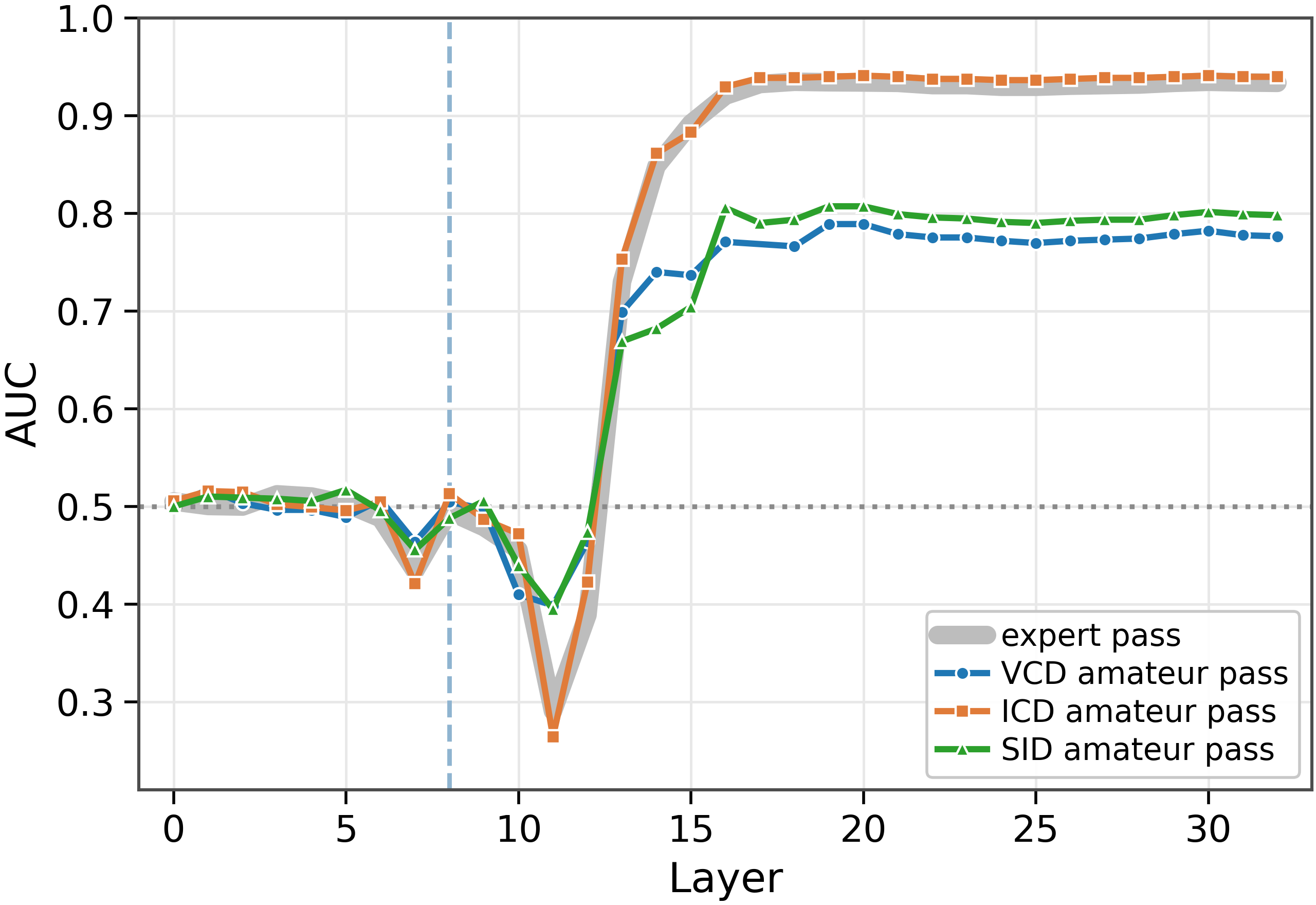}
  \hfill
  \includegraphics[width=0.48\linewidth]{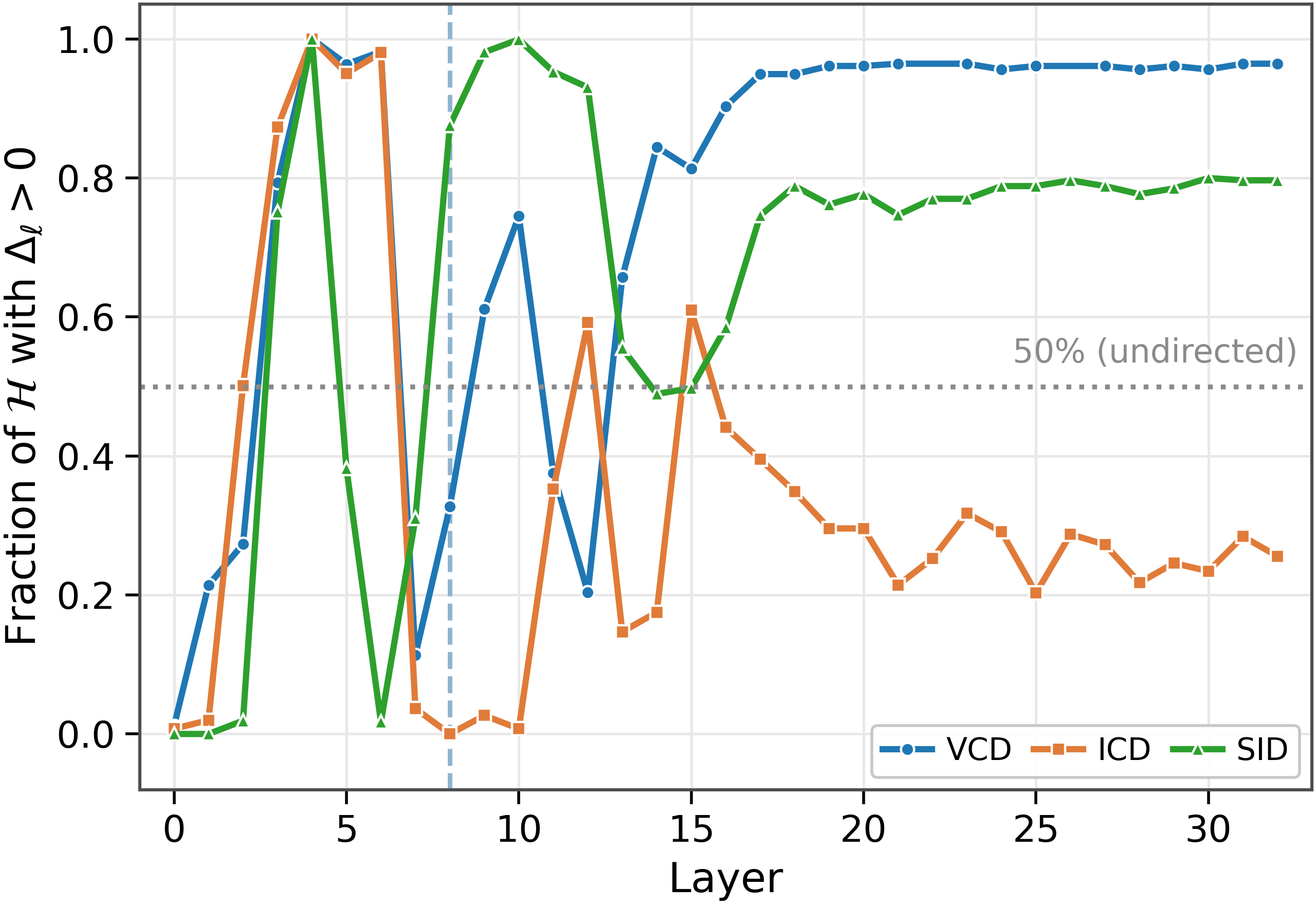}
  \caption{Layer-wise presence signal in the amateur pass and direction of the contrastive
  shift, LLaVA-v1.5-7B. \textbf{Left:} each image-corrupting amateur retains a degraded presence
  signal, while ICD's tracks the expert. \textbf{Right:} from layer 17 the sign of the shift is
  fixed by the amateur construction.}
  \label{fig:disc-battery-appx}
\end{figure}

A correction that targets hallucination should act once the model has enough information to tell
a hallucination apart from a correct answer. This subsection locates two depths and compares them.
The first is the depth at which that information becomes available, and the second is the depth at
which the contrastive shift is strongest.

Availability is measured with two linear probes fitted per layer on the expert-pass hidden state.
One predicts whether the queried object is present in the image, over all $9{,}000$ questions. The
other predicts whether the model hallucinates, among the questions whose ground-truth answer is
``No''. Both are logistic regressions with 5-fold cross-validation, and accuracy is reported as
out-of-fold AUC. Action is measured by the layer-wise mean of $\Delta_\ell$ over $\Hcal$.
Figure~\ref{fig:probe-vs-delta} in Section~\ref{sec:disc-results} plots the three curves together,
and Table~\ref{tab:app-depth} lists the values.

\begin{table}[!tb]
  \centering
  \small
  \setlength{\tabcolsep}{10pt}
  \renewcommand{\arraystretch}{1.15}
  \caption{Where the hallucination signal forms against where the contrastive shift acts,
  LLaVA-v1.5-7B with VCD. Probe values are AUC, shift values are log-odds.}
  \label{tab:app-depth}
  \begin{tabular}{@{}lcr@{}}
    \toprule
    Quantity & Layer & Value \\
    \midrule
    Object presence probe    & 8   & 0.958 \\
    Hallucination probe      & 12  & 0.970 \\
    \midrule
    Mean shift, layers 0--12 & --- & \phantom{$+$}$0.19$ \\
    Peak shift               & 19  & $+17.88$ \\
    Blocks 13--19            & --- & $+18.17$ \\
    Blocks 20--32            & --- & $-14.56$ \\
    \midrule
    Shift at output layer    & 32  & \phantom{$1$}$+3.32$ \\
    \bottomrule
  \end{tabular}
\end{table}

The two depths do not coincide. Object presence becomes decodable at layer 8 and hallucination
status at layer 12, both well before the answer is produced. Across those first twelve layers the
mean shift on $\Hcal$ is $0.19$ log-odds, which is negligible against the median expert margin of
$3.16$. The shift reaches its peak of $+17.88$ only at layer 19, seven blocks after the
hallucination is fully decodable.

\begin{figure}[!tb]
  \centering
  \includegraphics[width=0.86\linewidth]{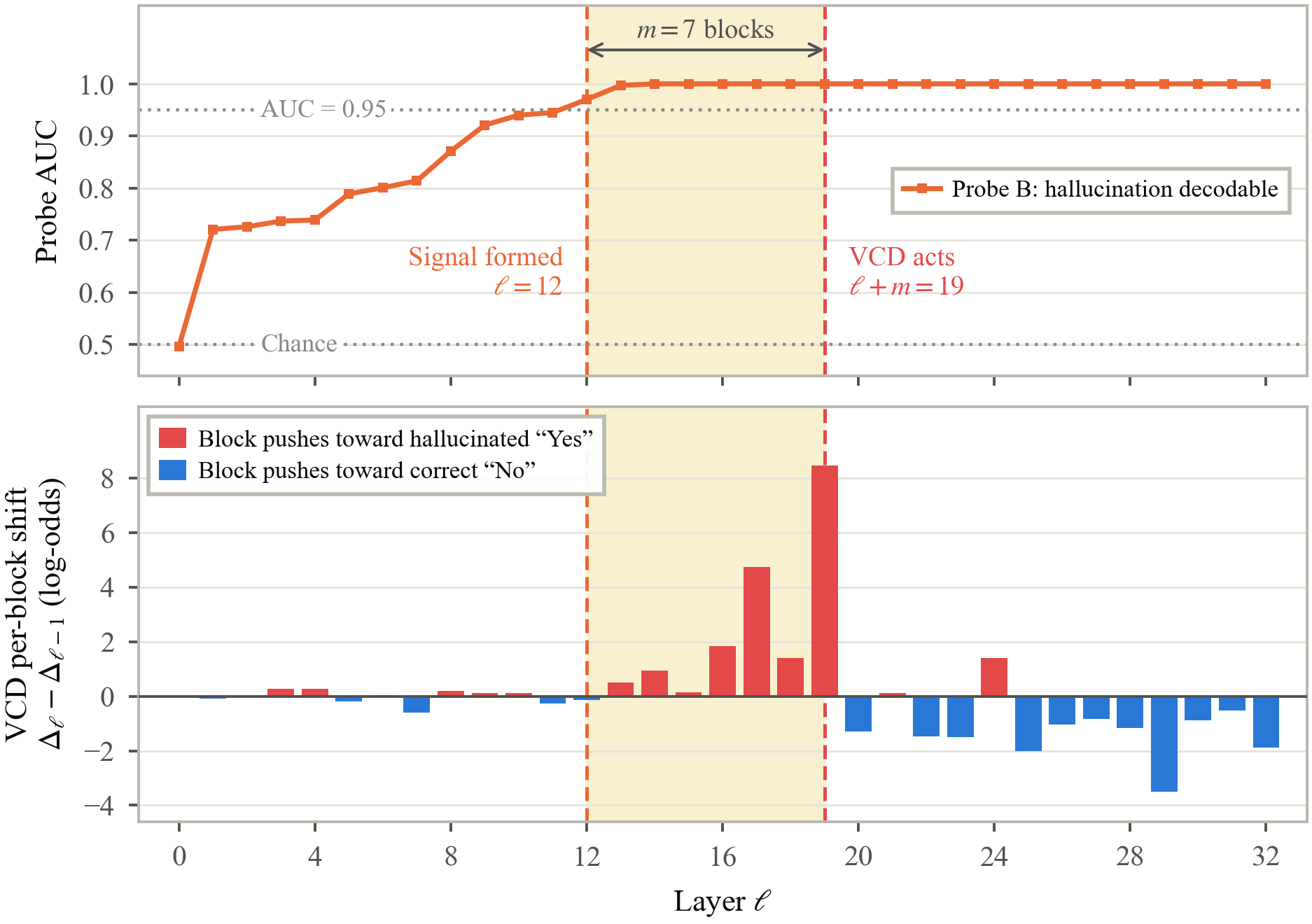}
  \caption{Per-block view of the same comparison, LLaVA-v1.5-7B with VCD. \textbf{Top:} the
  hallucination probe, fully formed at layer 12. \textbf{Bottom:} the per-block increment of the
  shift on $\Hcal$, red where a block pushes towards the hallucinated ``Yes''. The shaded window is
  the gap between the two.}
  \label{fig:app-exp1-blocks}
\end{figure}

Most of that peak is then undone by the network itself. Figure~\ref{fig:app-exp1-blocks}
decomposes the trajectory into per-block increments. Blocks 13--19 add $+18.17$ log-odds and blocks
20--32 remove $-14.56$, so $81.5\%$ of the peak is retracted and $+3.32$ survives to the output
layer. That residue is positive, which on a hallucination is the wrong direction.

This comparison is reported in the appendix rather than the main text because a depth gap alone is
not decisive. The residual stream is additive, so a shift written at layer 19 still reaches the
output, and a late shift is not for that reason an uninformed one. What the comparison establishes
is narrower. The information a selective correction would need is available for two thirds of the
network before the shift does anything, and the depth at which the shift is largest is not the
depth at which that information lives. Whether the shift is \emph{about} the hallucination at any
depth is the question Section~\ref{sec:disc-results} answers directly.

Figure~\ref{fig:disc-battery-appx} gives two further layer-wise views used in
Section~\ref{sec:disc-results}. The left panel shows how much of the presence signal each amateur
pass retains, and the right panel shows the depth beyond which the sign of the shift is fixed by
the amateur construction.

\subsection{Oracle Calibration}
\label{app:disc-oracle}

A selectivity AUC near $0.5$ admits two readings. Either the contrastive shift carries no
information about whether a question is a hallucination, or it carries some and the test is too
blunt to see it. The two controls below separate these cases by asking what the same test reports
on a shift that is selective by construction.

The first control fixes an upper bound. For each question we form an oracle shift
$\Delta_{\mathrm{oracle}}=-(1-p_{+})$, where $p_{+}$ is an out-of-fold probe estimate that the
queried object is present. This shift is large and negative exactly when the object is likely
absent, so it suppresses ``Yes'' on hallucinations by construction. Passing it through the same
selectivity computation gives the ceiling, the highest AUC the test can return on this data.

\begin{figure}[!tb]
  \centering

\includegraphics[width=0.82\linewidth]{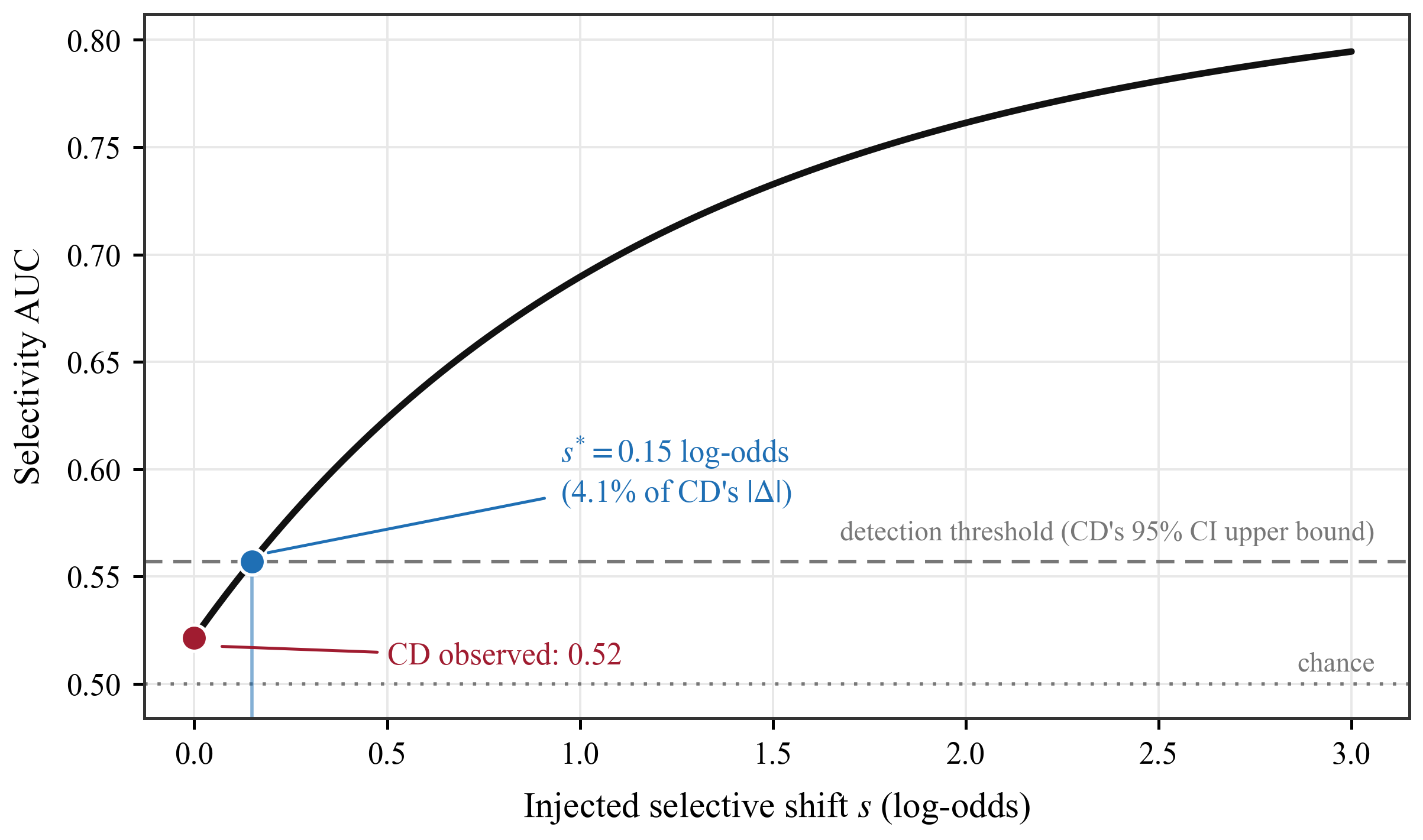}
  \caption{Detection limit of the selectivity test, VCD. The curve is measured selectivity as a
  selective component of size $s$ is injected into the method's own shift. The marker at $s=0$ is
  VCD as it actually runs and the dashed line is the detection threshold.}
  \label{fig:app-oracle}
\end{figure}

The second control fixes a lower bound. We add a selective component of size $s$ log-odds to each
method's own shift and increase $s$ until the measured selectivity clears the upper end of that
method's bootstrap interval. The smallest such $s^{*}$ is the detection limit, the weakest
selective behaviour the test would have registered. Dividing each method's mean $|\Delta|$ by
$s^{*}$ gives the detection factor, which states how much larger the observed shift is than the
smallest selective one the test could separate from noise. Figure~\ref{fig:app-oracle} shows the
sweep for VCD and Table~\ref{tab:app-oracle} reports both quantities for all three methods.

\begin{table}[!tb]
  \centering
  \small
  \setlength{\tabcolsep}{7pt}
  \renewcommand{\arraystretch}{1.15}
  \caption{Oracle calibration of the selectivity test, LLaVA-v1.5-7B. \emph{Ceiling} is the
  selectivity of the oracle shift, $s^{*}$ the detection limit, and \emph{Factor} the ratio of the
  mean $|\Delta|$ to $s^{*}$.}
  \label{tab:app-oracle}
  \begin{tabular}{@{}lcccccc@{}}
    \toprule
    \multirow{2.5}{*}{Method} & \multicolumn{3}{c}{Selectivity}
      & \multicolumn{3}{c}{Detection} \\
    \cmidrule(lr){2-4} \cmidrule(lr){5-7}
      & AUC & $95\%$ CI & Ceiling & Mean $|\Delta|$ & $s^{*}$ & Factor \\
    \midrule
    VCD & 0.523 & [0.49, 0.55] & 0.885 & 3.64 & 0.15 & $24.3\times$ \\
    ICD & 0.639 & [0.60, 0.67] & 0.889 & 0.18 & 0.02 & $\phantom{0}9.1\times$ \\
    SID & 0.691 & [0.66, 0.72] & 0.889 & 3.08 & 0.17 & $18.1\times$ \\
    \bottomrule
  \end{tabular}
\end{table}

Both bounds support the same reading. The ceiling in Table~\ref{tab:app-oracle} is
$0.885$--$0.889$, so the computation does return a high AUC when a selective shift is present. The
detection factors in the same table are between $9$ and $24$, so each method's shift is an order of
magnitude larger than the smallest selective one the test would have caught. A shift of that
magnitude carrying even a tenth of the oracle's hallucination-specificity would have registered.
The near-chance value for VCD is therefore a measurement of absence and not a failure of
resolution.

The two methods that do clear chance do so for reasons already visible in
Table~\ref{tab:disc-battery}. SID pushes both $\Hcal$ and $\Tcal$ towards ``Yes'' and merely pushes
$\Tcal$ further, so its ranking reflects a difference in degree rather than a correction. ICD
separates the two populations by $+0.11$ log-odds, which is small against the median expert margin
of $3.16$ and therefore cannot change an answer.

\subsection{Scalar-Bias Sweep}
\label{app:scalar-bias}

On a binary question the contrastive term contributes a single number to the Yes/No margin of each
sample. If that number carries hallucination-specific information, the information must lie in how
it varies across samples, since a quantity identical on every question can only move the decision
threshold. The final control therefore asks how much of the reported benefit survives once that
variation is removed.

\begin{figure}[!tb]
  \centering
  \includegraphics[width=0.82\linewidth]{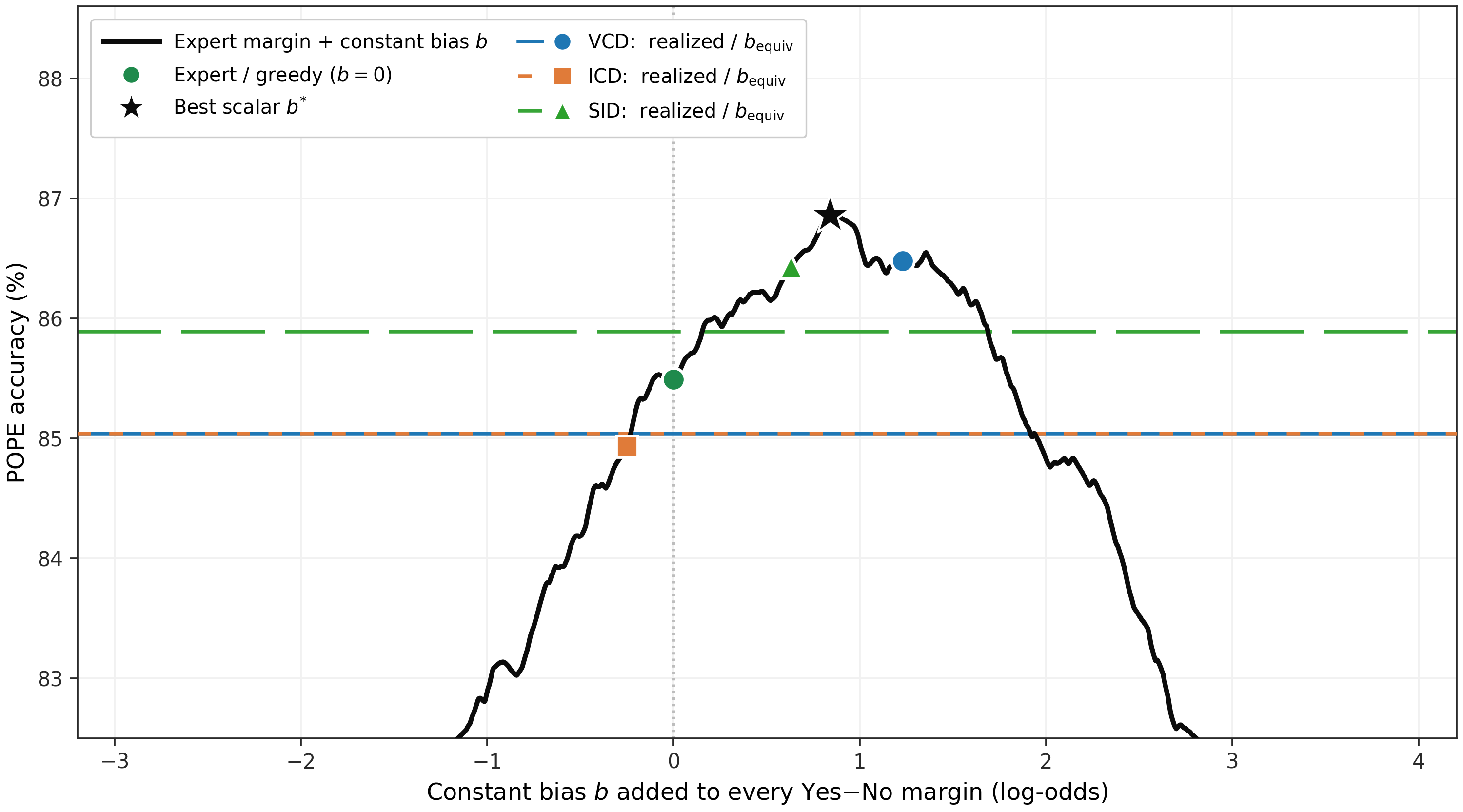}
  \caption{POPE accuracy under a single constant $b$ added to every Yes/No margin,
  LLaVA-v1.5-7B. Here $b=0$ is greedy decoding. Dashed lines mark each method's realised
  contrastive accuracy and markers its own mean shift flattened to a constant.}
  \label{fig:app-scalar}
\end{figure}

We replace the contrastive term with a constant $b$ added to the expert margin, so the predicted
answer becomes $[\,m^{E}_{32}+b>0\,]$, and sweep $b$ over $[-6,6]$ log-odds against POPE accuracy.
At $b=0$ the decision reduces to greedy decoding on the expert pass. Three points on this curve
matter: each method's realised contrastive accuracy, the constant $b_{\mathrm{equiv}}$ obtained by
flattening that method's own mean shift, and the accuracy-maximising constant $b^{*}$.
Figure~\ref{fig:app-scalar} shows the curve with these points marked and
Table~\ref{tab:app-scalar} lists them.

\begin{table}[!tb]
  \centering
  \small
  \setlength{\tabcolsep}{5pt}
  \renewcommand{\arraystretch}{1.15}
  \caption{Scalar-bias sweep, LLaVA-v1.5-7B on POPE-COCO. Accuracies and yes-rates are in
  percent, constants in log-odds. \emph{Greedy} is the expert decision at $b=0$.}
  \label{tab:app-scalar}
  \begin{tabular}{@{}lcccccccc@{}}
    \toprule
    \multirow{2.5}{*}{Method} & \multicolumn{2}{c}{Accuracy}
      & \multicolumn{2}{c}{Own mean $b_{\mathrm{equiv}}$}
      & \multicolumn{2}{c}{Best constant $b^{*}$}
      & \multicolumn{2}{c}{Yes-rate} \\
    \cmidrule(lr){2-3} \cmidrule(lr){4-5} \cmidrule(lr){6-7} \cmidrule(lr){8-9}
      & Greedy & Contrastive & $b$ & Accuracy & $b$ & Accuracy & Expert & CD \\
    \midrule
    VCD & 85.49 & 85.04 & $+1.23$ & 86.48 & $+0.84$ & 86.86 & 41.2 & 48.2 \\
    ICD & 85.47 & 85.04 & $-0.25$ & 84.93 & $+0.86$ & 86.89 & 41.3 & 40.0 \\
    SID & 85.47 & 85.89 & $+0.63$ & 86.43 & $+0.86$ & 86.89 & 41.3 & 44.3 \\
    \bottomrule
  \end{tabular}
\end{table}

A single tuned constant outperforms every contrastive method. Table~\ref{tab:app-scalar} shows that
$b^{*}=+0.84$ reaches $86.86\%$ accuracy, against $85.89\%$ for the best contrastive method and
$85.49\%$ for greedy decoding. Because $b^{*}$ is the same for every question, this accuracy is
obtained without any information about the image.

The comparison against $b_{\mathrm{equiv}}$ is tighter, since it holds each method's average effect
fixed and removes only the per-sample variation. Doing so \emph{improves} accuracy on two of the
three methods, VCD from $85.04$ to $86.48$ and SID from $85.89$ to $86.43$, and costs ICD $0.11$
points. The per-sample structure of the contrastive term therefore contributes nothing on this
benchmark, and on two methods it is harmful.

This is what a change of decision threshold looks like from the outside. A constant cannot
distinguish one question from another, so a method that is matched or beaten by its own mean is
doing nothing that a threshold shift could not do. The near-zero $R^{2}$ in the right panel of
Figure~\ref{fig:disc-battery} is the same finding measured directly on $\Delta_\ell$. The yes-rate
column of Table~\ref{tab:app-scalar} states it in POPE's own units. Every method moves the
operating point, VCD by 7 points and SID by 3, yet only SID ends above the expert it is applied to,
by $+0.42$, while VCD and ICD end $0.44$ and $0.42$ points below it.

\section{Jaccard Similarity Analysis}
\label{app:jaccard}
\subsection{Similarity Metric}
\label{app:jaccard-metric}
Section~\ref{sec:apc-greedy} measures how close the output of one decoding condition is to the
output of another. The two responses are free-form text of different lengths, so we compare the
vocabulary they use rather than matching them position by position. The Jaccard index provides
this comparison.

For two finite sets $A$ and $B$,
\begin{equation}
\jacc(A,B) \;=\; \frac{|A \cap B|}{|A \cup B|} \;\in\; [0,1],
\qquad \jacc(\varnothing,\varnothing) \triangleq 1,
\label{eq:jaccard}
\end{equation}
that is, the number of elements common to both sets divided by the number of distinct elements
occurring in either. The index equals $1$ when the two sets are identical and $0$ when they are
disjoint, and the stated convention treats two empty responses as identical.

We apply the index to text by mapping a response $y$ to the set $T(y)$ of unique sub-word token
identifiers assigned to it by the model's own tokenizer. The similarity between two responses
$y_A$ and $y_B$ is then $\jacc(T(y_A), T(y_B))$. Since $T(y)$ discards order and repetition, this
measures the overlap in what a condition says and not the order in which it says it.
Figure~\ref{fig:apc-both-refs} reports the resulting similarities and
Table~\ref{tab:apc-both-refs-selected-with-CI} gives them with confidence intervals.

\subsection{Confidence Intervals}
\label{sec:appendix-C-CI-Results}

Table~\ref{tab:apc-both-refs-selected-with-CI} gives the numbers plotted in
Figure~\ref{fig:apc-both-refs}, together with paired bootstrap $95\%$ confidence intervals. For
every value of $\beta$, and on all three models, APC-S is closer to greedy decoding than to direct
sampling. The lowest and highest values of $\beta$ are special cases, and we explain each below.

\begin{table}[H]
\centering
\caption{Jaccard similarity of APC-S to greedy decoding and to direct sampling across $\beta$,
on LLaVA-Bench (In-the-Wild). Paired bootstrap $95\%$ confidence intervals in brackets.}
\label{tab:apc-both-refs-selected-with-CI}
\small
\setlength{\tabcolsep}{2pt}
\renewcommand{\arraystretch}{1.15}
\begin{tabular}{@{}lcccccc@{}}
\toprule
\multirow{2.5}{*}{$\beta$} & \multicolumn{2}{c}{LLaVA-v1.5-7B} & \multicolumn{2}{c}{Qwen2.5-VL-7B-Instruct}
& \multicolumn{2}{c}{InternVL3-8B} \\
\cmidrule(lr){2-3} \cmidrule(lr){4-5} \cmidrule(lr){6-7}
& Greedy & Sampling & Greedy & Sampling & Greedy & Sampling \\
\midrule
0 & 0.339 {\footnotesize [.293,\,.391]} & 0.279 {\footnotesize [.239,\,.323]}
    & 0.371 {\footnotesize [.328,\,.422]} & 0.317 {\footnotesize [.276,\,.363]}
    & 0.399 {\footnotesize [.348,\,.455]} & 0.342 {\footnotesize [.298,\,.392]} \\
0.1 & 0.405 {\footnotesize [.352,\,.465]} & 0.314 {\footnotesize [.273,\,.360]}
    & 0.428 {\footnotesize [.377,\,.484]} & 0.346 {\footnotesize [.302,\,.397]}
    & 0.420 {\footnotesize [.378,\,.466]} & 0.357 {\footnotesize [.316,\,.405]} \\
0.2 & 0.405 {\footnotesize [.357,\,.450]} & 0.312 {\footnotesize [.271,\,.357]}
    & 0.477 {\footnotesize [.424,\,.534]} & 0.371 {\footnotesize [.320,\,.430]}
    & 0.475 {\footnotesize [.418,\,.534]} & 0.415 {\footnotesize [.357,\,.479]} \\
0.4 & 0.455 {\footnotesize [.403,\,.511]} & 0.333 {\footnotesize [.291,\,.379]}
    & 0.496 {\footnotesize [.443,\,.554]} & 0.369 {\footnotesize [.322,\,.421]}
    & 0.516 {\footnotesize [.460,\,.575]} & 0.419 {\footnotesize [.362,\,.482]} \\
0.6 & 0.507 {\footnotesize [.458,\,.559]} & 0.337 {\footnotesize [.294,\,.385]}
    & 0.552 {\footnotesize [.494,\,.613]} & 0.383 {\footnotesize [.334,\,.439]}
    & 0.564 {\footnotesize [.507,\,.624]} & 0.405 {\footnotesize [.352,\,.464]} \\
0.8 & 0.583 {\footnotesize [.526,\,.641]} & 0.342 {\footnotesize [.299,\,.390]}
    & 0.620 {\footnotesize [.557,\,.682]} & 0.379 {\footnotesize [.330,\,.436]}
    & 0.628 {\footnotesize [.567,\,.692]} & 0.417 {\footnotesize [.363,\,.477]} \\
\textbf{1.0} & \textbf{0.968} {\footnotesize\bfseries [.936,\,.993]} & \textbf{0.351} {\footnotesize\bfseries [.305,\,.403]}
& \textbf{0.779} {\footnotesize\bfseries [.721,\,.835]} & \textbf{0.378} {\footnotesize\bfseries [.330,\,.435]}
& \textbf{0.733} {\footnotesize\bfseries [.674,\,.790]} & \textbf{0.415} {\footnotesize\bfseries [.360,\,.476]} \\
\bottomrule
\end{tabular}
\end{table}

At $\beta=0$ the adaptive plausibility constraint is inactive. The cutoff removes no candidate, so
the output is plain sampling. This row is reported as a reference point rather than as an APC
setting.

At $\beta=1$ the cutoff sits at the highest expert logit, so only the top candidate should survive
at each step and the output should reproduce greedy decoding exactly. The measured similarity is
high but below one. Logits are floating-point values, and the top candidate can lie arbitrarily
close to a tied or near-tied second choice, in which case the constrained sampler may select a
different token at that step. Because generation is autoregressive, one such divergence early in a
response changes the continuation, even though the constraint behaves almost identically to greedy
decoding at every step.

\subsection{Robustness Checks}
\label{app:jaccard-checks}

The unigram index of Section~\ref{app:jaccard-metric} ignores word order and counts every token
equally. A high similarity could therefore arise from shared function words alone, or from the
same words arranged differently. We repeat the measurement under two stricter definitions of the
token set. The \emph{bigram} variant forms the set from adjacent token pairs, which makes it
sensitive to order. The \emph{content-word} variant removes common English stopwords before
forming the set, so only content-bearing words contribute.

Table~\ref{tab:corroborate} reports both variants at the two ends of the sweep. The pattern found
under the unigram index holds in each case. Similarity to greedy decoding is low at $\beta=0$ and
close to one at $\beta=1$, on both models. The convergence towards greedy decoding therefore
reflects agreement in content and in order, and is not an artefact of the metric.

\begin{table}[H]
\centering
\caption{Jaccard similarity of APC-S to greedy decoding at the two ends of the $\beta$ sweep,
under three definitions of the token set. Paired bootstrap $95\%$ confidence intervals in
brackets.}
\label{tab:corroborate}
\small
\setlength{\tabcolsep}{6pt}
\renewcommand{\arraystretch}{1.2}
\begin{tabular}{@{}llcc@{}}
\toprule
\multirow{2.5}{*}{Model} & \multirow{2.5}{*}{Token set}
  & \multicolumn{2}{c}{Similarity to greedy decoding} \\
\cmidrule(lr){3-4}
  & & $\beta=0$ & $\beta=1$ \\
\midrule
\multirow{3}{*}{LLaVA-v1.5-7B}
  & Unigram      & 0.3371 {\footnotesize [0.2912, 0.3900]} & 0.9683 {\footnotesize [0.9363, 0.9931]} \\
  & Bigram       & 0.1866 {\footnotesize [0.1384, 0.2451]} & 0.9577 {\footnotesize [0.9150, 0.9902]} \\
  & Content-word & 0.2692 {\footnotesize [0.2216, 0.3250]} & 0.9629 {\footnotesize [0.9256, 0.9919]} \\
\midrule
\multirow{3}{*}{Qwen2.5-VL-7B-Instruct}
  & Unigram      & 0.3713 {\footnotesize [0.3264, 0.4221]} & 0.7785 {\footnotesize [0.7210, 0.8344]} \\
  & Bigram       & 0.1968 {\footnotesize [0.1473, 0.2562]} & 0.6958 {\footnotesize [0.6176, 0.7717]} \\
  & Content-word & 0.3330 {\footnotesize [0.2806, 0.3934]} & 0.7520 {\footnotesize [0.6877, 0.8150]} \\
\bottomrule
\end{tabular}
\end{table}

\section{Contrastive Adjustment on Object Tokens}
\label{top10_section}

\subsection{Top-10 Candidate Population}

\begin{table}[h]
  \centering
  \small
  \setlength{\tabcolsep}{6pt}
  \renewcommand{\arraystretch}{1.15}
  \caption{Effect of the contrastive adjustment $d_t$ on real and hallucinated object tokens in
  the top-10 candidate population. Amp.\ and Sup.\ are the percentages of tokens amplified
  ($d_t>0$) and suppressed ($d_t<0$); the larger of the two is shown in bold.}
  \label{tab:top10_table}
  \begin{tabular}{@{}llrrrrrr@{}}
    \toprule
    \multirow{2.5}{*}{Model} & \multirow{2.5}{*}{Method}
      & \multicolumn{3}{c}{Real} & \multicolumn{3}{c}{Hallucinated} \\
    \cmidrule(lr){3-5} \cmidrule(lr){6-8}
      & & Mean $d_t$ & Amp.\% \ & Sup.\% \ & Mean $d_t$ & Amp.\% \ & Sup.\%\ \\
    \midrule
    \multirow{3}{*}{LLaVA-v1.5-7B}
      & VCD & $0.061$  & \textbf{53.3} & 46.7 & $0.016$  & 49.7 & \textbf{50.3} \\
      & SID & $0.505$  & \textbf{58.7} & 41.3 & $-0.193$ & 38.8 & \textbf{61.2} \\
      & ICD & $0.087$  & \textbf{59.3} & 40.7 & $0.089$  & \textbf{63.8} & 36.2 \\
    \midrule
    \multirow{3}{*}{Qwen2.5-VL-7B-Instruct}
      & VCD & $-0.327$ & 37.6 & \textbf{62.4} & $-0.250$ & 39.0 & \textbf{61.0} \\
      & SID & $0.565$  & \textbf{58.2} & 41.8 & $0.225$  & \textbf{51.9} & 48.1 \\
      & ICD & $0.105$  & 49.9 & \textbf{50.1} & $0.094$  & 49.8 & \textbf{50.2} \\
    \midrule
    \multirow{3}{*}{InternVL3-8B}
      & VCD & $-0.027$ & 45.0 & \textbf{55.0} & $-0.018$ & 45.9 & \textbf{54.1} \\
      & SID & $1.084$  & \textbf{66.0} & 34.0 & $0.428$  & \textbf{56.2} & 43.8 \\
      & ICD & $0.013$  & 46.0 & \textbf{54.0} & $0.063$  & 49.7 & \textbf{50.3} \\
    \bottomrule
  \end{tabular}
\end{table}

Section~\ref{sec:no-signal} examined the object tokens the model actually emitted. This
subsection widens the population to the ten highest-scoring candidates at every generation step,
so that the conclusion does not depend on which token was finally selected. For each model and
method, Table~\ref{tab:top10_table} reports the mean contrastive adjustment $d_t$ together with
the share of tokens it amplifies and the share it suppresses, separately for real and for
hallucinated object tokens.

A method that selectively suppresses hallucination should amplify real object tokens and suppress
hallucinated ones, so the two halves of the table should point in opposite directions. In seven of
the nine model-method pairs they point the same way, with the majority of both classes either
amplified or suppressed together. Only LLaVA-v1.5-7B with SID shows the intended pattern, where
$58.7\%$ of real tokens are amplified and $61.2\%$ of hallucinated tokens are suppressed. The same
model with VCD is nominally in the right direction, but both margins sit within a few points of
chance and support no conclusion.

The wider population therefore reproduces the finding of Table~\ref{tab:emitted}. The contrastive
adjustment moves real and hallucinated object tokens in the same direction in almost every case,
so it does not distinguish between them.

\subsection{Motivation for the Gaussian Proxy}

\begin{figure}[h]
  \centering
  \includegraphics[width=\linewidth]{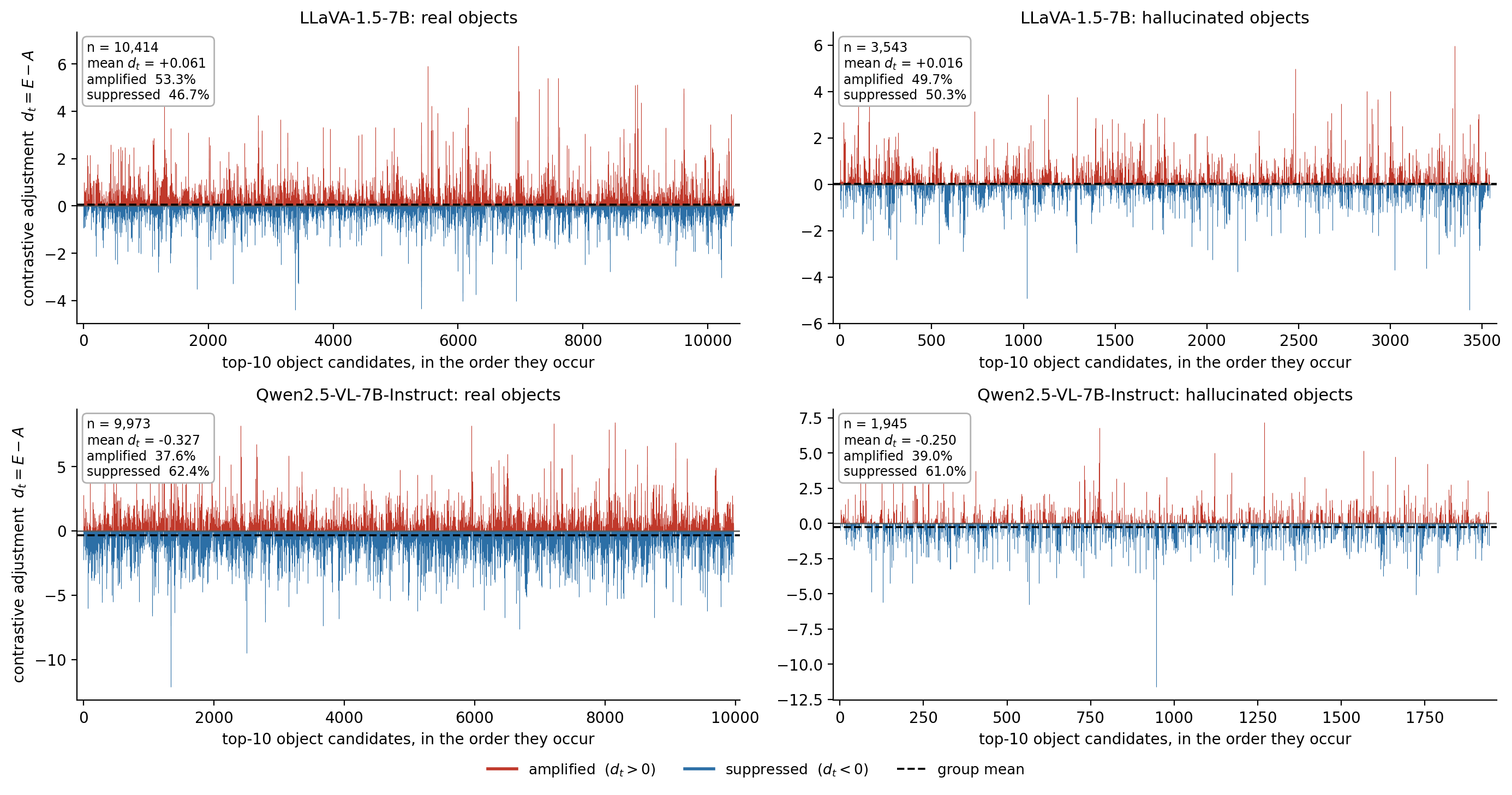}
  \caption{Contrastive adjustment $d_t$ for each top-10 object candidate under VCD. Columns:
  real and hallucinated objects. Rows: LLaVA-v1.5-7B and Qwen2.5-VL-7B-Instruct.}
  \label{fig:d-vcd}
\end{figure}

Figure~\ref{fig:d-vcd} plots $d_t$ for every top-10 object candidate under VCD, separately for
real and for hallucinated objects. The traces show no trend across the sequence and no visible
separation between the two classes. The values fluctuate about a mean close to zero, in the manner
of noise rather than of a signal that targets particular tokens.

This observation motivated the control reported in Table~\ref{tab:proxy-chair-perrun}. If $d_t$
behaves like noise, then replacing it with actual noise of matched mean and variance should not
degrade generation quality. The Gaussian proxy of Section~\ref{sec:no-signal} implements that
substitution, and its CHAIR scores test whether the structure of $d_t$ carries anything the
matched noise does not.

\clearpage
\input{checklist}

\end{document}

%% file: checklist.tex
\section*{NeurIPS Paper Checklist}

\begin{enumerate}

\item {\bf Claims}
    \item[] Question: Do the main claims made in the abstract and introduction accurately reflect the paper's contributions and scope?
    \item[] Answer: \answerYes{} 
    \item[] Justification: We have supported the claims and contributions in the abstract and introduction in the paper.

\item {\bf Limitations}
    \item[] Question: Does the paper discuss the limitations of the work performed by the authors?
    \item[] Answer: \answerYes{} 
    \item[] Justification: We have discussed the limitations in the "Limitations and Future Work" section

\item {\bf Theory assumptions and proofs}
    \item[] Question: For each theoretical result, does the paper provide the full set of assumptions and a complete (and correct) proof?
    \item[] Answer: \answerNA{} 
    \item[] Justification: The paper does not include any theoretical results.

    \item {\bf Experimental result reproducibility}
    \item[] Question: Does the paper fully disclose all the information needed to reproduce the main experimental results of the paper to the extent that it affects the main claims and/or conclusions of the paper (regardless of whether the code and data are provided or not)?
    \item[] Answer: \answerYes{} 
    \item[] Justification: We have provided explanations of methodologies and hyperparameter values used in the main content and Appendix respectively.

\item {\bf Open access to data and code}
    \item[] Question: Does the paper provide open access to the data and code, with sufficient instructions to faithfully reproduce the main experimental results, as described in supplemental material?
    \item[] Answer: \answerNo{} 
    \item[] Justification: We will provide the source code for experiments once accepted.

\item {\bf Experimental setting/details}
    \item[] Question: Does the paper specify all the training and test details (e.g., data splits, hyperparameters, how they were chosen, type of optimizer) necessary to understand the results?
    \item[] Answer: \answerYes{} 
    \item[] Justification: We have included the datasets used in the main content and all additional experimental details in the appendix 

\item {\bf Experiment statistical significance}
    \item[] Question: Does the paper report error bars suitably and correctly defined or other appropriate information about the statistical significance of the experiments?
    \item[] Answer: \answerYes{} 
    \item[] Justification: We have defined and provided information about confidence intervals used and runs on multiple seeds

\item {\bf Experiments compute resources}
    \item[] Question: For each experiment, does the paper provide sufficient information on the computer resources (type of compute workers, memory, time of execution) needed to reproduce the experiments?
    \item[] Answer: \answerNo{} 
    \item[] Justification: We have provided the GPU used in the Appendix. We will provide other relevant information related on acceptance.
    
\item {\bf Code of ethics}
    \item[] Question: Does the research conducted in the paper conform, in every respect, with the NeurIPS Code of Ethics \url{https://neurips.cc/public/EthicsGuidelines}?
    \item[] Answer: \answerYes{} 
    \item[] Justification: We reviewed and followed the NeurIPS Code of Ethics.

\item {\bf Broader impacts}
    \item[] Question: Does the paper discuss both potential positive societal impacts and negative societal impacts of the work performed?
    \item[] Answer: \answerYes{} 
    \item[] Justification: We have discussed the broader impacts in the Appendix.
    
\item {\bf Safeguards}
    \item[] Question: Does the paper describe safeguards that have been put in place for responsible release of data or models that have a high risk for misuse (e.g., pre-trained language models, image generators, or scraped datasets)?
    \item[] Answer: \answerNA{} 
    \item[] Justification: We have only used public datasets and models. The paper poses no such risks.

\item {\bf Licenses for existing assets}
    \item[] Question: Are the creators or original owners of assets (e.g., code, data, models), used in the paper, properly credited and are the license and terms of use explicitly mentioned and properly respected?
    \item[] Answer: \answerYes{} 
    \item[] Justification: All creators and original owners of assets have been cited in the "References" Section.

\item {\bf New assets}
    \item[] Question: Are new assets introduced in the paper well documented and is the documentation provided alongside the assets?
    \item[] Answer: \answerNo{} 
    \item[] Justification: The paper does not release new datasets or models. Codes used for experiments will be provided on acceptance.

\item {\bf Crowdsourcing and research with human subjects}
    \item[] Question: For crowdsourcing experiments and research with human subjects, does the paper include the full text of instructions given to participants and screenshots, if applicable, as well as details about compensation (if any)? 
    \item[] Answer: \answerNA{} 
    \item[] Justification: The paper does not involve crowdsourcing nor research with human subjects.

\item {\bf Institutional review board (IRB) approvals or equivalent for research with human subjects}
    \item[] Question: Does the paper describe potential risks incurred by study participants, whether such risks were disclosed to the subjects, and whether Institutional Review Board (IRB) approvals (or an equivalent approval/review based on the requirements of your country or institution) were obtained?
    \item[] Answer: \answerNA{} 
    \item[] Justification: The paper does not involve crowdsourcing nor research with human subjects.

\item {\bf Declaration of LLM usage}
    \item[] Question: Does the paper describe the usage of LLMs if it is an important, original, or non-standard component of the core methods in this research? Note that if the LLM is used only for writing, editing, or formatting purposes and does \emph{not} impact the core methodology, scientific rigor, or originality of the research, declaration is not required.
    \item[] Answer: \answerNA{} 
    \item[] Justification: LLMs have only been used for formatting and grammatical purposes.

\end{enumerate}